\documentclass[10pt,twocolumn,letterpaper]{article}

\usepackage[pagenumbers]{cvpr} 

\definecolor{cvprblue}{rgb}{0.21,0.49,0.74}
\usepackage[pagebackref,breaklinks,colorlinks,allcolors=cvprblue]{hyperref}
\usepackage{booktabs}
\usepackage{colortbl}
\usepackage{multirow}

\def\paperID{*****} 
\def\confName{CVPR}
\def\confYear{2026}

\title{Implicit Neural Representation Facilitates Unified Universal Vision Encoding}

\author{Matthew Gwilliam \quad Xiao Wang \quad Xuefeng Hu \quad Zhenheng Yang\\
  TikTok*
}

\begin{document}
\maketitle
\begin{abstract}
Models for image representation learning are typically designed for either recognition or generation.
Various forms of contrastive learning help models learn to convert images to embeddings that are useful for classification, detection, and segmentation.
On the other hand, models can be trained to reconstruct images with pixel-wise, perceptual, and adversarial losses in order to learn a latent space that is useful for image generation.
We seek to unify these two directions with a first-of-its-kind model that learns representations which are simultaneously useful for recognition and generation.
We train our model as a hyper-network for implicit neural representation, which learns to map images to model weights for fast, accurate reconstruction.
We further integrate our INR hyper-network with knowledge distillation to improve its generalization and performance.
Beyond the novel training design, the model also learns an unprecedented compressed embedding space with outstanding performance for various visual tasks.
The complete model competes with state-of-the-art results for image representation learning, while also enabling generative capabilities with its high-quality tiny embeddings.
The code is available at \href{https://github.com/tiktok/huvr}{this https URL}.
\iftoggle{cvprfinal}{
\begingroup
\renewcommand\thefootnote{\fnsymbol{footnote}}
\footnotetext[1]{This work is for research purposes only and is not currently integrated into any TikTok technology.}
\endgroup
}{}
\end{abstract}    
\section{Introduction}
\label{sec:introduction}

\begin{figure}[h]
\begin{center}
\includegraphics[width=\linewidth]{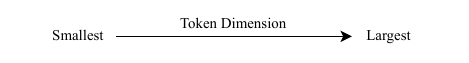}\vspace{-1em}
\includegraphics[width=\linewidth]{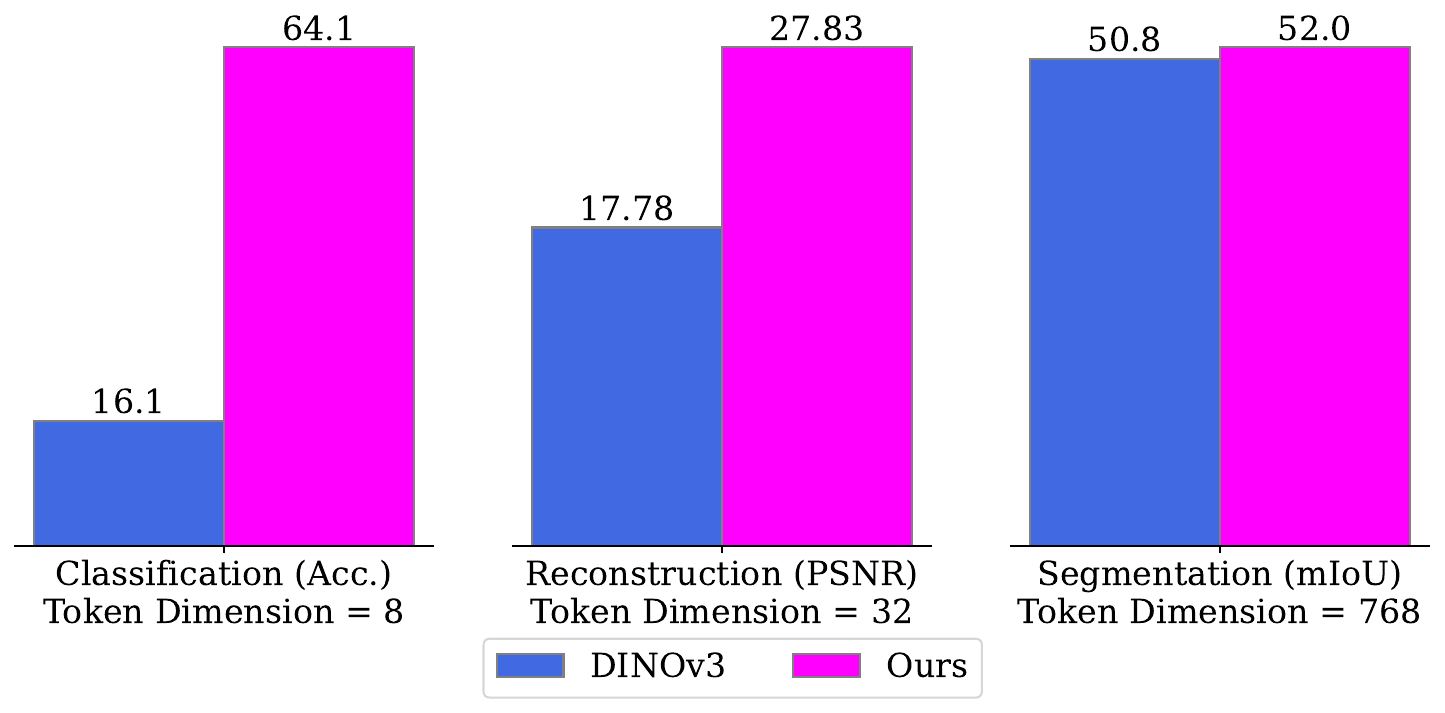}
\includegraphics[width=\linewidth]{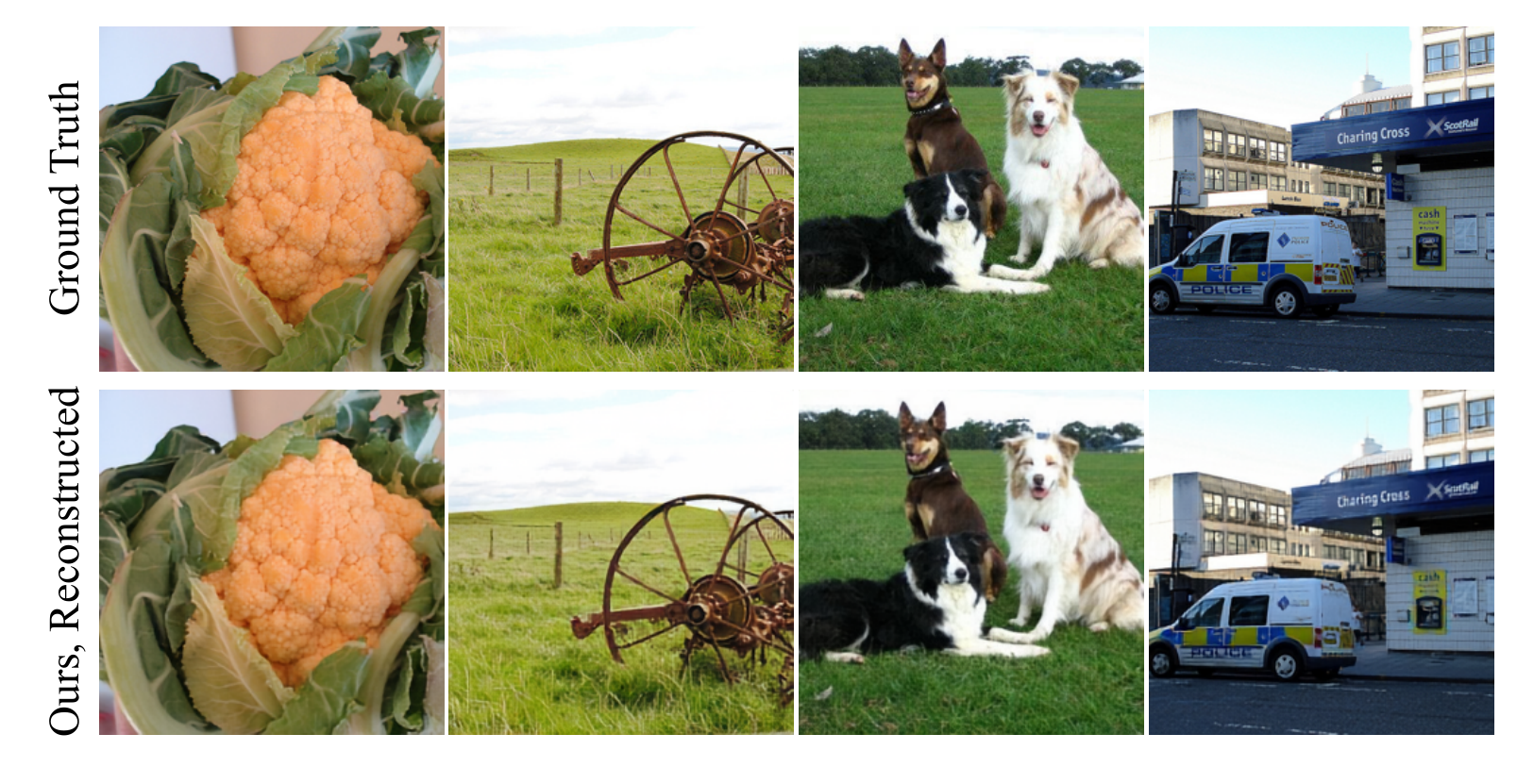} 
\caption{\textbf{(Top)} Unified modeling, in terms of both tasks and token dimensions. We propose implicit neural \textbf{H}yper-networks for \textbf{U}nified \textbf{V}isual \textbf{R}epresentation, \textbf{HUVR}, with good \textbf{classification}, \textbf{reconstruction}, and \textbf{segmentation} (shown for ViT-B/16 on ImageNet, ImageNet, and ADE20K, respectively). We design our model to generate not only standard-sized tokens, but also \textbf{Tin}y \textbf{Tok}en\textbf{s} (TinToks). Here, the tiny embeddings of DINOv3 are generated via principle component analysis (PCA). \textbf{(Bottom)} Reconstruction. We unify recognition and generative task families.}
\label{fig:teaser}
\end{center}
\end{figure}

Vision encoders, which learn to compute useful image and video representations, are foundational components of modern computer vision systems. 
Although such encoders may be able to solve many different tasks, in practice, we typically use them in a specialized manner according to their strengths.
Models trained with text-image contrastive learning~\cite{radford2021learning} lend themselves well to tasks such as retrieval and visual question answering.
Models trained with image-only contrastive learning~\cite{chen2020simple,misra2020self} or self-distillation~\cite{caron2021emerging,oquab2023dinov2} perform well for fine-grained image understanding, such as semantic segmentation.
Variational autoencoders, which reconstruct pixels, enable image and video synthesis with diffusion and auto-regressive models.
Some prior and concurrent works attempt to unify these recognition-focused and generative-focused models post-hoc~\cite{donahue2019large,li2022mage,mukhopadhyay2024text,zheng2025diffusiontransformersrepresentationautoencoders}.
These works point to a very promising synergy between pretraining methods for recognition and generative models.
A natively unified encoder's features should have good high level (image classification), mid level (semantic segmentation), low level (depth estimation), and pixel level (reconstruction) information out-of-the-box.

These requirements point to hyper-networks for implicit neural representation (INR)~\cite{chen2022transformers,kim2022generalizable} as a natural candidate. 
An INR hyper-network is a network that takes image inputs, and predicts neural network weights (INR) as output.
These output INRs take pixel coordinates as inputs, and give pixel color channel values as outputs.
INR hyper-networks are similar to traditional autoencoders in the sense that their latents have good pixel representation. 
However, unlike convolutional VAEs, transformer-based INR hyper-networks can easily borrow encoder design, patch sizes, and latent dimensions from state-of-the-art Vision Transformer (ViT) encoders~\cite{dosovitskiy2020image}.
Furthermore, compared to other representation learning methods, INR hyper-networks perform compression along multiple axes, first by converting a specific image to latents, and second by learning a shared ``base'' INR to represent all possible images. 
We hypothesize this is helpful for learning high quality embeddings not only at the pixel-level, but also for low-, middle-, and high-level image information~\cite{solomonoff1964formal,schmidhuber2006computer,lee2021compressivevisualrepresentations,gregor2016conceptualcompression}.

Unified representation requires more than simply training a model whose representations have suitable semantics for any task.
In practice, tasks differ from each other not just in terms of the level of information or semantics they need, but also in terms of the amount of computational resources available to solve them.
Tasks like retrieval become very difficult as the amount of data increases at scale, and reducing the embedding size introduces massive savings.
Thus, we propose to unify representations not just in terms of tasks, but also by designing a model that can generate both compressed and non-compressed representations.

We design our ``\textbf{H}yper-network for \textbf{U}nified \textbf{V}isual \textbf{R}epresentation'' (HUVR) with a novel INR hyper-network design to natively unify disparate representation learning families, particularly recognition and reconstruction, in terms of both architectural design and pretraining.
HUVR encodes images both in standard size representations, as well as a smaller representation that we refer to as ``\textbf{Tin}y \textbf{Tok}ens'' (TinToks). 
Figure~\ref{fig:teaser} shows their superior recognition and novel reconstruction abilities.

We build on the strengths of INR hyper-networks to design HUVR and its TinToks.
INR hyper-networks do not natively learn good high level semantics, and the design of the latent tokens makes it difficult to recover patch-level information.
We solve the gap in semantics via knowledge distillation from a pretrained vision encoder.
We refactor the design of the latent tokens to naturally support both a mapping between input patches and patch tokens, as well as a global ``cls'' token.
To create the compressed representation, the TinToks, we introduce learnable feature downsampling and upsampling layers in between the transformer backbone and the layers that produce the final INR prediction.

In summary, our key contributions are:


\begin{itemize}
    \item We design and train an INR hyper-network, HUVR, which can match or outperform DINOv3~\cite{simeoni2025dinov3} -- with ViT-B/16 we achieve +0.4\% for ImageNet~\cite{imagenet} classification and +1.2 mIoU for ADE20K~\cite{zhou2017scene} semantic segmentation, and +4.84 PSNR for reconstruction.
    \item The compressed representation TinToks, at 96x compression, can offer +48\% ImageNet classification compared to a DINOv3 PCA baseline, and +1.26 PSNR compared to the Stable Diffusion VAE~\cite{rombach2022highresolutionimagesynthesislatent} at equal embedding size.
    \item An improved design for hyper-networks for image implicit neural representations yields +2.34 PSNR even with only 10\% training time on ImageNette. 
\end{itemize}

\section{Related Work}
\label{sec:related_work}

\subsection{Implicit Neural Representation}

\noindent\textbf{Implicit Neural Representation.}
An implicit neural representation (INR) is a neural network that maps coordinates to a signal (image, video, 3D, audio, etc.)~\cite{sitzmann2020implicit,tancik2020fourfeat,M_ller_2022,mildenhall2020nerf,chen2021nerv,xu2022signal,saragadam2023wire}.
Due to their small size, a significant volume of research focuses on trying to leverage INRs for image and video compression~\cite{dupont2021coin,dupont2022coin++,strümpler2022implicitneuralrepresentationsimage,ladune2023coolchiccoordinatebasedlowcomplexity,chen2021nerv,zhang2021implicitneuralvideocompression,kim2022scalable,maiya2023nirvana,kwan2024nvrcneuralvideorepresentation,kim2023c3highperformancelowcomplexityneural}.
Methods in the NeRV~\cite{chen2021nerv} family use a combination of linear layers, convolutional layers, and PixelShuffle~\cite{shi2016realtimesingleimagevideo} operations to map frame embeddings to the RGB values of the frame for video compression~\cite{li2022enervexpediteneuralvideo,Zhang_2024_CVPR,Yan_2024_CVPR,Kim_2024,kwan2023hinerv,Lee_2023,chen2023hnerv,chen2022cnerv,He_2023_CVPR,Zhao_2023_CVPR,Saethre_2024_CVPR,Zhao_2024_CVPR,xu2024vqnervvectorquantizedneural,wu2024qs,gwilliam2025design}.
Other INR methods (including some in the NeRV family) use grid features for positional embedding~\cite{Lee_2023,kwan2023hinerv,kim2022scalable,maiya2023nirvana,girish2023shacirascalablehashgridcompression}.

\noindent\textbf{Hyper-Networks.}
INRs require costly per-sample training, which limits their applicability. 
Some try to avoid the need to train per-sample by instead training a hyper-network.
The hyper-network learns to predict INR network weights that can reconstruct the input~\cite{chen2022transformers,kim2022generalizable,lee2023localityawaregeneralizableimplicitneural,chen2024fastencodingdecodingimplicit,zhang2024attention} or generate novel outputs~\cite{skorokhodov2021adversarialgenerationcontinuousimages,Haydarov_2024_CVPR,yu2022generatingvideosdynamicsawareimplicit}.
Latent-INR redesigns INRs as \textit{per-video} hyper-networks, and aligns an learnable latent for each frame with its corresponding CLIP embedding~\cite{maiya2024latent}.
While we also pursue INR with semantics, unlike Latent-INR, we learn a general hyper-network that can predict INR network weights even for inputs not seen during training.

\begin{figure*}[h]
\begin{center}
\includegraphics[width=\linewidth]{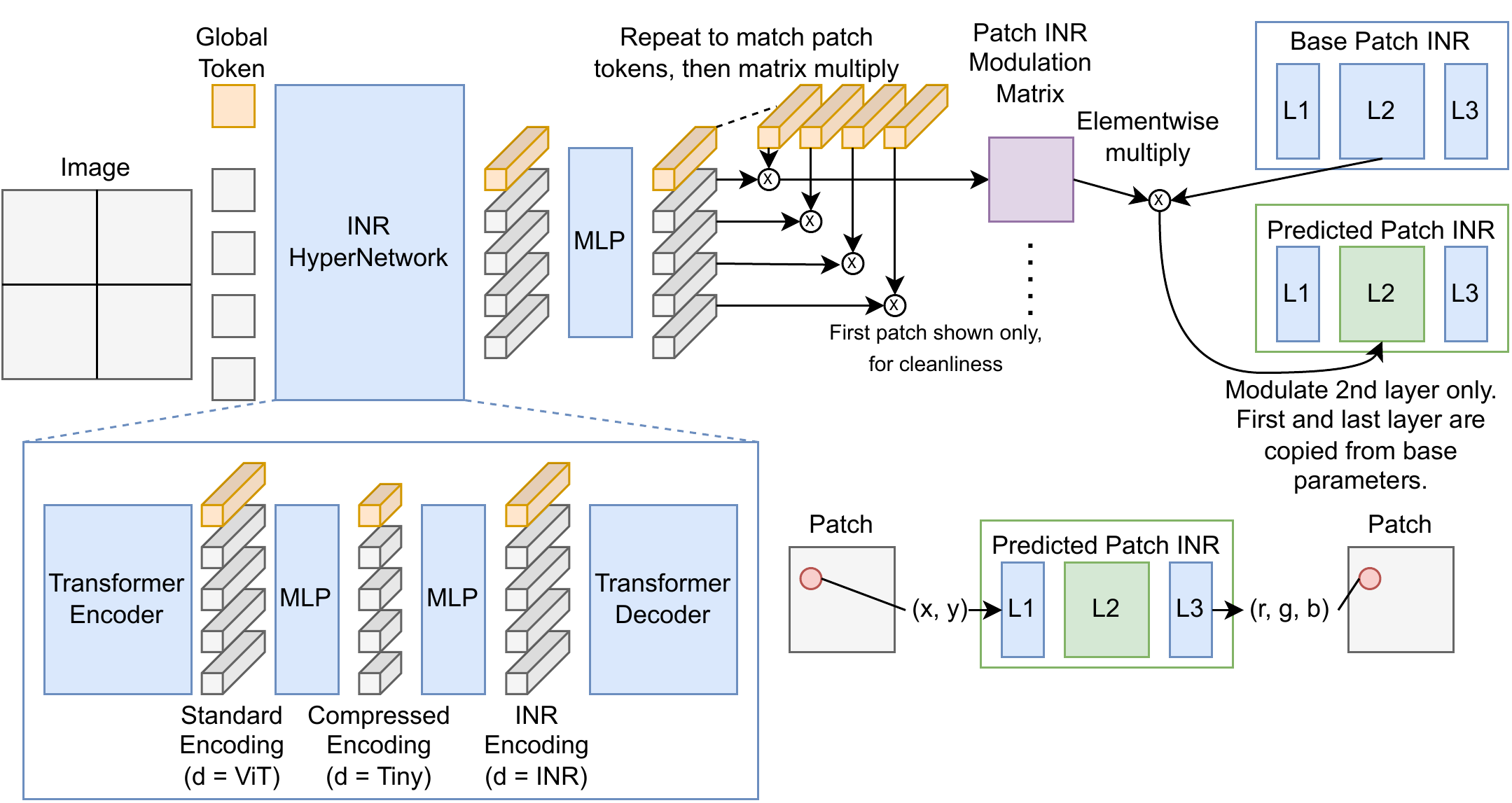}
\caption{INR Hyper-Network for unified visual representation. The standard and compressed encodings from our model have powerful recognition and reconstruction capabilities, enabling downstream tasks ranging from classification to image generation.}
\label{fig:method}
\end{center}
\end{figure*}

\subsection{Image Representation Learning}
The first unsupervised deep learning methods for image representation focused on restoring withheld or corrupted information~\cite{zhang2016colorful,noroozi2016unsupervised,misra2020self,pathak2016context}, but they struggled to match the performance of supervised learning.
These were followed by methods, mainly based on contrastive learning and clustering, which were able to compete with supervised learning~\cite{chen2020simple,chen2020big,DBLP:journals/corr/abs-2003-04297,chen2021mocov3,pmlr-v139-zbontar21a,tomasev2022pushing,DBLP:journals/corr/abs-2006-07733,chen2021exploring,Bardes2021VICRegVR,caron2018deep,caron2020unsupervised,caron2021emerging,pang2022unsupervised,zhou2022mugs,li2022efficient}.
iBOT~\cite{zhou2022ibot} was able to outperform these with masked image modeling~\cite{DBLP:journals/corr/abs-2111-06377,assran2022masked,bao2022beit,huang2022contrastive}, and recent entries in the DINO family fuse these with a self-distillation paradigm based on knowledge distillation~\cite{hinton2015distilling} for state-of-the-art results~\cite{oquab2023dinov2,simeoni2025dinov3}.
Some methods train an image representation jointly with text modeling~\cite{radford2021learning,zhai2023sigmoid,Ranzinger_2024_CVPR,heinrich2025radiov25improvedbaselinesagglomerative,sun2024eva}, and these methods typically perform best as the vision encoders for multimodal large language models~\cite{bai2023qwen,wang2024qwen2,liu2023visual,liu2024improved,xue2024xgen}, while text-free methods (DINOv3) perform best for dense tasks.
Some methods train models for both recognition and generation~\cite{li2022mage,Xiang_2023_ICCV,donahue2016adversarial,dumoulin2016adversarially,donahue2019large,chen2016infogan,nie2020semi}, while other recent works rely on a pre-trained generative model, which is then adapted post-hoc for recognition~\cite{mukhopadhyay2024text,Xiang_2023_ICCV,li2023your}.
In this work, we propose a text-free method with a design that allows it to natively perform recognition and reconstruction.
\section{Methods}
\label{sec:method}

\subsection{Hyper-Network for Implicit Neural Representation}
\label{subsec:hyper-networks}

\noindent\textbf{Background.} Implicit neural representations store signals as neural network weights. 
A function $f_\theta : X \rightarrow I$, which maps coordinates $X$ to an image (or more generally, a signal) $I$.
In the case of images, we can represent these coordinates as $X=\{(x_i,y_j | 0 \leq x_i < W, 0 \leq y_j < H \}$ and images as color tuples $I=\{(r_{x_i,y_j},g_{x_i,y_j},b_{x_i,y_j})|0 \leq x_i < W, 0 \leq y_j < H \}$.
In practice, these networks require long per-sample training time, with multiple iterations for every pixel, for $f_\theta$ to ``memorize'' $I$.
Additionally, there is no `generalizability' -- for every new $I$, we must train a new corresponding $f_\theta$.

To solve this, a hyper-network approximates a function, $h$, for mapping input signals to their corresponding instance-specific INR, $h: I \rightarrow f_{\theta'}$~\cite{chen2022transformers}.
Instead of a fitting $f_\theta$ separately for each sample, $h$ is trained on a large dataset, and learns a general mapping between data samples and $\theta'$.
After this pre-training, encoding $I$ as $\theta'$ requires only one forward pass of $h$.

Prior INR hyper-networks are initialized with 4 learnable components: (i) a transformer encoder, $E$, (ii) ``shared'' or ``base'' INR weights $\theta_b$ with $n$ layers, $\theta_b = \{W_i\}_{i=1}^{n}$, (iii) input weight tokens corresponding to the number of layers $n$, and (iv) fully-connected (FC) linear layers to project between the transformer encoder output dimension and $d_\text{in}$ of $W_i$.
The transformer encoder, $E$, takes the image and weight tokens as input, and gives output image and weight tokens.
The output image tokens are discarded.
The output weight tokens corresponding to some $W_i$ in $\theta_b$ are fed through an FC layer to change their token dimension to $d_\text{in}$, and then repeated $k=\frac{d_\text{out}}{n}$ times to form a modulation matrix $M_i$ of shape $d_\text{in}{\times}d_\text{out}$.
Each $M_i$ is multiplied elementwise with $W_i$ to modulate it, forming the final weights $\theta'$ of $f$.
In other words, for a sample-specific $f_{\theta'}$, we have $\theta' = \{W_i \odot M_i \}_{i=1}^{n}$.
For more details, we include a tutorial in the Appendix.



\noindent\textbf{Overview.} Figure~\ref{fig:method} depicts our method.
Unlike prior works, there are no learnable weight tokens.
Instead of predicting image INRs, our $h$ predicts $f_{\theta'}$ for each patch separately.
Patch tokens are fed to the INR hyper-network, along with a novel learnable global token.
The INR hyper-network itself is composed of 3 parts, which generate 3 sets of tokens in sequence: (i) standard Vision Transformer (ViT) output, (ii) compressed encoding (TinToks), (iii) an INR modulation encoding.
The INR modulation encoding for each patch is projected and then multiplied by a projection of the global token, which yields a matrix $M_i$ to modulate the corresponding $W_i$ of $\theta_b$.
We use all modulated, per-patch INRs for a given image to reconstruct the input patch.
We concatenate the reconstructed patches to reconstruct the full $I$.

\noindent\textbf{Key Innovation \#1: Patch Tokens as Weight Tokens.}
Existing INR hyper-networks discard the output image tokens, using them neither to calculate loss nor perform inference.
This makes dense tasks like semantic segmentation very challenging, since $I$ is represented mainly in the weight tokens, but these lack clear correlation with spatial locations in the input.
To resolve the inefficiency and lack of capability, we use the data tokens themselves as weight tokens.
This presents a challenge -- in the standard hyper-network formulation, the number of weight tokens must be a factor of the $d_\text{in}$ or $d_\text{out}$ for $W_i$ in $\theta_b$.
This makes it difficult to formulate good hyper-network and INR settings.
To solve this, we refactor the framework, predicting instead an $f_{\theta'}$ per-patch, rather than per-image.
So, the output tokens of $h$ are also patch tokens.

This leads to another issue.
The output token has dimension $d_\text{ViT}$, but $W_i$ has dimension $d_\text{in}{\times}d_\text{out}$.
Following the ideas from prior works~\cite{chen2022transformers,chen2024fastencodingdecodingimplicit} we could simply make sure $d_\text{ViT} = d_\text{in}$ and, and copy the output token $d_\text{out}$ times to yield an $M_i$ to multiply with $W_i$.
However, in practice this is suboptimal (see Section~\ref{subsec:inr_results}).

\noindent\textbf{Key Innovation \#2: Global Tokens to Modulate and Summarize.} 
Now, recall that the original INR hyper-network formulation does not have a cls token.
We solve both of our problems at once.
We add a global token $g$, which can act as a cls token for recognition tasks.
We can project $g$ to have dimension $d_g=d_\text{out}$ and project patch token $p$ to have dimension $d_p=d_\text{in}$.
Then, when we multiply $g \times p^T$, we yield the matrix $M_i$ with shape $d_\text{in}{\times}d_\text{out}$ for modulating $W_i$.
With this, we successfully modify the hyper-network architecture to have a class token, patch tokens, and no wasted tokens, making it suitable for both INR prediction and image recognition.

\noindent\textbf{Key Innovation \#3: Tiny tokens (TinToks).} 
In practice, we want to be able to set the dimension of the transformer encoder, $d_\text{ViT}$ separately from $d_\text{in}$ and $d_\text{out}$ for any $W_i$.
Additionally, compute-constrained applications require smaller tokens.
To satisfy this, we introduce an intermediate representation, TinToks, which have their own dimension $d_t$.
To accomplish this, we use a linear layer to downsample from $d_\text{ViT}$ to $d_t$, and another to upsample for decoding.
We then process the tokens with a transformer decoder to allow for better reconstruction.
We use final linear projections to convert the tokens from the decoding dimension to $d_\text{in}$ and $d_\text{out}$ for the patch and global tokens, respectively.

\noindent\textbf{Training.} We train the model with distillation losses (described in Section~\ref{subsec:distillation}) and visual quality losses.
For the visual quality, we compute mean-squared error on pixels between the input image and the concatenated patch predictions of each patch $f_{\theta'}$.
We can also optionally train with SSIM~\cite{wang2004image} and LPIPS~\cite{zhang2018perceptual} losses to further improve the reconstruction.

\subsection{Distillation to INR Hyper-Networks}
\label{subsec:distillation}



\noindent\textbf{Key Innovation \#4: Distillation for Unified Reprsentation.} We align our representations to the pre-trained model (for most of our experiments, DINOv3) by computing a loss between DINOv3 features and a linear projection of our features.
While we could hypothetically apply this to the outputs of any of ours layers, in practice we apply this loss to the outputs of the last encoder (enc) and decoder (dec) blocks.
We learn separate transformations for each block, and within each block we learn separate transformations depending on whether a token is global ($\text{type}_g$ or patch ($\text{type}_p$.
So, considering token types ${\text{type}_g, \text{type}_p} \in T$ and outputs of the final encoder and decoder blocks ${o_\text{enc}, o_\text{dec}} \in O$, we get 4 components of our loss, which we weight individually with some $\alpha$.
We can compute an $L_{\text{2}}$ distillation loss with features $F$ from some teacher as follows:

\begin{equation}
\label{eq:distillation}
    L_{distillation} = \sum_{t}^{T} \sum_{o}^{O} L_{\text{2}}(\theta_{t,o}(F_{t,o}) - F_{t,\text{teacher}}) * \alpha_{t,o}  
\end{equation}

Note that we do not perform distillation on the compressed tokens directly.
Combined with the INR reconstruction objective, we find that distillation to the encoder and decoder is sufficient to imbue the compressed tokens with good semantics for downstream reconstruction tasks.
\section{Results}
\label{sec:results}

\begin{table*}[t!]
	\centering
	\caption{\textbf{Tiny token results.} We compare our compressed tokens to a principle component analysis (PCA) baseline. We compute the PCA transforms using the ImageNet-1k training set. We report linear probing classification accuracies for ImageNet (original and ReaL labels), ObjectNet, and some fine-grained datasets. We train a decoder on frozen DINOv3 for a reconstruction baseline.}
	\resizebox{1.0\linewidth}{!}{
		\begin{tabular}{@{}lcl cccccccc cc@{}}
			\toprule
			\multicolumn{3}{c}{Settings} &
			\multicolumn{8}{c}{Classification} &
            \multicolumn{2}{c}{Reconstruction} \\
			\cmidrule{1-3}
			\cmidrule(l){4-11}
			\cmidrule(l){12-13}
    Enc. & Dim. & Pre-Training & INet & ReaL & ONet & Cars & CUB & DTD & Flowers & Food & PSNR & SSIM \\ 
    \midrule
    VAE & - & SD VAE & - & - & - & - & - & - & - & - & 24.99 & 0.7078 \\ 
    \midrule
    \cellcolor{white}    & \cellcolor{white}    & DINOv3, PCA & 16.1 & 18.0 & 8.2 & 31.3 & 44.9 & 28.8 & 57.8 & 31.7 & 15.51 & 0.4975 \\
    \cellcolor{white}    & \multirow{-2}{*}{8} \cellcolor{white}    & \textbf{HUVR (ours)} & \textbf{64.1} & \textbf{71.1} & \textbf{34.0} & \textbf{53.8} & \textbf{74.8} & \textbf{39.6} & \textbf{98.9} & \textbf{70.5} & \textbf{24.66} & \textbf{0.6918} \\ 
    \cmidrule{2-13}
    \cellcolor{white}    & \cellcolor{white}    & DINOv3, PCA & 40.6 & 45.1 & 19.5 & 58.5 & 67.6 & 45.5 & 86.7 & 59.2 & 16.75 & 0.5228 \\
    \cellcolor{white}    & \multirow{-2}{*}{16} \cellcolor{white}    & \textbf{HUVR (ours)} & \textbf{75.6} & \textbf{82.6} & \textbf{43.7} & \textbf{72.5} & \textbf{83.1} & \textbf{59.8} & \textbf{99.4} & \textbf{85.8} & \textbf{26.25} & \textbf{0.7361} \\ 
    \cmidrule{2-13}
    \cellcolor{white}    & \cellcolor{white}    & C-RADIOv3, PCA & \underline{71.8} & \underline{79.2} & \underline{42.1} & 73.9 & 64.6 & \textbf{71.0} & 93.7 & 83.3 & - & - \\
    \cellcolor{white}    & \cellcolor{white}    & SigLIP 2, PCA & 62.3 & 67.8 & 31.3 & \textbf{86.8} & 75.4 & \underline{70.2} & \underline{98.6} & \underline{85.9} & - & - \\
    \cellcolor{white}    & \cellcolor{white}    & DINOv3, PCA & 64.1 & 70.3 & 35.7 & 80.0 & 80.7 & 64.5 & 97.3 & 78.9 & 17.68 & 0.5398 \\
    \multirow{-8}{*}{ViT-B} \cellcolor{white}    & \multirow{-4}{*}{32} \cellcolor{white}    & \textbf{HUVR (ours)} & \textbf{79.4} & \textbf{85.6} & \textbf{47.0} & \underline{80.8} & \textbf{86.5} & 70.1 & \textbf{99.6} & \textbf{89.7} & \textbf{27.83} & \textbf{0.7845} \\ 
    \midrule
    \cellcolor{white}    & \cellcolor{white}    & C-RADIOv3, PCA & \underline{77.9} & \textbf{83.8} & \underline{47.5} & 82.0 & 73.2 & \textbf{73.3} & 98.6 & \underline{89.4} & - & - \\
    \cellcolor{white}    & \cellcolor{white}    & SigLIP 2, PCA & 63.4 & 67.9 & 33.9 & \textbf{90.6} & 77.8 & \underline{67.0} & \underline{99.3} & 89.2 & - & - \\
    \cellcolor{white}    & \cellcolor{white}    & DINOv3, PCA & 72.2 & 77.5 & 45.0 & 84.2 & \textbf{84.2} & 64.8 & 99.1 & 88.1 & 17.27 & 0.5311 \\
    \multirow{-4}{*}{ViT-L} \cellcolor{white}    & \multirow{-4}{*}{32} \cellcolor{white}    & 
    \textbf{HUVR (ours)} & \textbf{78.1} & \underline{82.9} & \textbf{53.9} & \underline{88.1} & \underline{81.4} & 66.8 & \textbf{99.5} & \textbf{91.2} & \textbf{27.70} & \textbf{0.7802} \\ 
    \bottomrule
    \end{tabular}
    }
    \label{tab:tiny_token_results}
\end{table*}

\begin{table}[t!]
	\centering
	\caption{\textbf{Diffusion results.} We train a class-conditional DiT-XL on ImageNet using the original VAE latents, and other DiT-XLs using our compressed tokens. We compare the DiTs in terms of FID, Inception Score, Precision and Recall.}
	\resizebox{1.0\linewidth}{!}{
		\begin{tabular}{@{}ll ccccc@{}}
			\toprule
            Autoencoder & Latents & FID$\downarrow$ & sFID$\downarrow$ & IS$\uparrow$ & P$\uparrow$ & R$\uparrow$ \\
            \midrule
            SD VAE         & $32{\times}32{\times}4$ & \textbf{23.05} & \underline{68.65} & \textbf{70.34} & \textbf{0.4318} & \textbf{0.4775} \\
            \midrule
            Ours & $16{\times}16{\times}16$ & 24.72 & 76.09 & 60.17 & 0.3850 & \underline{0.4645} \\
            Ours & $16{\times}256{\times}256$ & \underline{24.53} & \textbf{68.37} & \underline{66.13} & \underline{0.4307} & 0.4367 \\
            \bottomrule
        \end{tabular}
    }
    \label{tab:diffusion_results}
\end{table}

\subsection{Settings}

\noindent\textbf{Training.} 
For our models, we use a modern version of the Vision Transformer (ViT)~\cite{dosovitskiy2020image}, with Rotary Positional Embeddings (RoPE)~\cite{su2024roformer}.
We focus on ViT-B/16 and ViT-L/16, due to the prevalence of practical adoption for these sizes, and to reduce our training cost.
Unless otherwise indicated, we pre-train our models on a mix of DataComp~\cite{datacomp} (image-only) and ImageNet22k~\cite{imagenet15russakovsky} data (without labels). 
We ensure at least 10\% of all samples are from ImageNet22k (inspired by DINOv3~\cite{simeoni2025dinov3}) for the equivalent of 50 epochs on ImageNet22k.
We parallelize according to resource availability by adjusting batch size and scaling the learning rate by a linear factor.
For data augmentation, we only use random resized cropping.
For details on model and training settings, see the Appendix.


\noindent\textbf{Evaluation.} 
Since we are mainly concerned with feature quality, we compute recognition evaluations with linear probing classification on ImageNet1k~\cite{imagenet} using both the original labels and ReaL labels~\cite{beyer2020imagenet}, and ObjectNet~\cite{barbu2019objectnet}.
We do not perform any finetuning.
We also perform linear probing with L-BFGS~\cite{liu1989limited} for 5 FGVC datasets -- Caltech-UCSD Birds 200~\cite{WahCUB_200_2011}, Stanford Cars~\cite{KrauseStarkDengFei-Fei_3DRR2013}, Describable Textures Dataset~\cite{cimpoi14describing}, Oxford 102 Flower~\cite{Nilsback08}, and Food-101~\cite{bossard14}.
We perform linear probing on ADE20K~\cite{zhou2017scene} for semantic segmentation following the setup from AM-RADIO~\cite{Ranzinger_2024_CVPR} and use a similar setup for a depth probe~\cite{El_Banani_2024_CVPR} on NYUv2~\cite{Silberman:ECCV12}.
For reconstruction, we report PSNR, SSIM, and LPIPS on the 50,000 images from the ImageNet-1k validation set (which is never seen during training). 
For generation, we follow the protocol from DiT~\cite{peebles2023scalablediffusionmodelstransformers}.
We also perform some comparison to prior INR hyper-networks by training and evaluating our reconstruction capability on comparable settings for the ImageNette subset of ImageNet~\cite{imagenette}, Celeb-A~\cite{liu2015faceattributes}, and LSUN Churches~\cite{yu2015lsun} to prove that HUVR is the state-of-the-art INR hyper-network for images.
For more details, see the Appendix.

\begin{table*}[t!]
	\centering
	\caption{\textbf{Standard classification results.} Our model is competitive with state-of-the-arts for classification, including on fine-grained classification (FGVC) datasets. We set our TinTok $d=32$ but evaluate the performance of the full, standard-sized tokens.}
	\resizebox{1.0\linewidth}{!}{
		\begin{tabular}{@{}ll cccccccc@{}}
			\toprule
			\multicolumn{2}{c}{Settings} &
			\multicolumn{8}{c}{Classification} \\
			\cmidrule{1-2}
			\cmidrule(l){3-10}
    Encoder & Pre-Training & INet & ReaL & ONet & Cars & CUB & DTD & Flowers & Food \\ 
    \midrule    
    \cellcolor{white}     & C-RADIOv3 & 82.4 & 87.6 & 54.5 & 88.6 & 79.8 & 83.5 & 99.1 & 91.5 \\ 
    
    \cellcolor{white}     & SigLIP 2 & 84.5 & \underline{89.0} & \textbf{68.4} & 92.8 & 82.6 & 83.7 & 99.3 & \underline{94.2} \\ 
    
    \cellcolor{white}    & DINOv3 & \underline{84.6} & 88.9 & 59.4 & \textbf{93.4} & \textbf{89.7} & \underline{83.9} & \textbf{99.7} & 93.7 \\ 
    \multirow{-4}{*}{ViT-B, Patch Size 16, $d=768$} \cellcolor{white}    & \textbf{HUVR (ours)} & \textbf{85.0} & \textbf{89.2} & \underline{62.0} & \underline{93.1} & \underline{89.4} & \textbf{84.3} & \textbf{99.7} & \textbf{94.3} \\ 
    \midrule
    
    \cellcolor{white}     & C-RADIOv3 & 86.0 & 89.6 & 66.0 & 92.8 & 85.4 & \underline{86.2} & 99.6 & 94.4 \\ 
    
    \cellcolor{white}     & SigLIP 2 & \textbf{87.1} & \textbf{90.1} & \textbf{75.5} & \textbf{94.9} & 85.0 & 85.1 & 99.6 & \textbf{96.1} \\ 
    
    \cellcolor{white}    & DINOv3 & \textbf{87.1} & 90.0 & \underline{71.0} & \underline{93.7} & \textbf{91.1} & \textbf{86.6} & \textbf{99.7} & \underline{95.6} \\ 
    \multirow{-4}{*}{ViT-L, Patch size 16, $d=1024$} \cellcolor{white}    & \textbf{HUVR (ours)} & 86.9 & \textbf{90.1} & 70.7 & 91.7 & \underline{85.7} & 84.5 & \textbf{99.7} & 95.4 \\ 
    \bottomrule
    \end{tabular}
    }
    \label{tab:main_results}
\end{table*}

\subsection{Tiny Tokens are a Unified Representation}
\label{subsec:tiny_tokens}

We show in Table~\ref{tab:tiny_token_results} that our compressed vector representations simultaneously achieve good recognition and reconstruction.
To our knowledge, our method is the first to attempt classification, segmentation, and depth estimation with such compressed features.
So, for a baseline, we fit a principle component analysis (PCA) transform for DINOv3~\cite{simeoni2025dinov3}, C-RADIOv3~\cite{heinrich2025radiov25improvedbaselinesagglomerative}, and SigLIP2~\cite{tschannen2025siglip2multilingualvisionlanguage} at the corresponding compressed token dimension on the training set of ImageNet1k.
We then use this transform for ImageNet, and all other datasets, and refer to it as ``PCA'' for each method.
Note that none of these methods perform reconstruction natively, but as a baseline we train a decoder with roughly the same number of parameters as ours, on the frozen DINOv3 PCA features.
Our compressed tokens offer the best unified modeling capability.
The gap between ours and the baseline increases as the compression ratio increases -- notice the difference between ours and DINOv3 for 8-dim tokens.

To show the promising generative potential of our latents, we train a DiT~\citep{peebles2023scalablediffusionmodelstransformers} on the latents of our model instead of the SD VAE~\cite{rombach2022highresolutionimagesynthesislatent} latents.
We show results in Table~\ref{tab:diffusion_results}.
While we fail to beat the performance of the existing SD VAE or achieve generative state-of-the-art results, we consider these results promising and hope they highlight the potential our method as a unified representation. 
Following insights from concurrent work~\cite{zheng2025diffusiontransformersrepresentationautoencoders}, we try increasing the TinTok dimension to 256, and observe that increasing the TinTok dimension seems beneficial for generation.
Applying further insights from the concurrent work, which uses frozen pre-trained vision encoders, would likely help further improve the performance, especially given that our model is natively superior for reconstruction.

\subsection{HUVR beats DINOv3 for Classification}
\label{subsec:standard_recognition}

The purpose of HUVR is not simply to generate powerful TinToks.
Even HUVR's standard-size embeddings are quite useful.
Table~\ref{tab:main_results} shows that HUVR can match or beat prior works on ImageNet (both original and ReaL labels), and achieve good results on ObjectNet and the fine-grained datasets as well. 
Comparing our ObjectNet to SigLIP 2, it is worth keeping in mind that SigLIP 2 trains on 25 times more data.
For FGVC  datasets, DINOv3 mines pretraining data that is similar to many of these, whereas we use pre-training data that is not specifically curated for these~\cite{datacomp}.

\subsection{HUVR can Perform Dense Recogntion Tasks}

\begin{table}[t!]
	\centering
	\caption{\textbf{Dense recognition results.} We compare our distilled hyper-networks to Radiov3, SigLIP 2, and DINOv3 for semantic segmentation (ADE20K) and depth estimation (NYUv2).}
	\resizebox{1.0\linewidth}{!}{
		\begin{tabular}{@{}lll ccc@{}}
			\toprule
			& & &
			\multicolumn{2}{c}{ADE20K} &
            \multicolumn{1}{c}{NYUv2} \\
			\cmidrule(l){4-5}
            \cmidrule(l){6-6}
    Model & Size & Method & mIoU & mAcc & RMSE$\downarrow$ \\
    \midrule
    &  & SigLIP 2 & 40.0 & 53.6 & 0.5317 \\
    &  & C-RADIOv3 & 49.5 & 62.4 & 0.3359 \\
    &  & DINOv3 & \underline{50.8} & \underline{62.6} & \underline{0.3305} \\
    & \multirow{-4}{*}{768} & \textbf{HUVR (ours)} & \textbf{52.0} & \textbf{63.4} & \textbf{0.3263} \\
    \cmidrule{2-6}
    &  & SigLIP 2 & 17.5 & 25.0 & 0.8155 \\
    &  & C-RADIOv3 & 28.9 & \textbf{40.3} & \underline{0.6017} \\
    &  & DINOv3 & \textbf{29.7} & 38.6 & 0.7056 \\  
    \multirow{-8}{*}{ViT-B} & \multirow{-4}{*}{32} & \textbf{HUVR (ours)} & \textbf{29.7} & \underline{39.1} & \textbf{0.5980} \\ 
    \midrule
    &  & SigLIP 2 & 42.0 & 55.8 & 0.5107 \\
    &  & C-RADIOv3 & 51.6 & 63.8 & \underline{0.3239} \\
    &  & DINOv3 & \textbf{54.2} & \textbf{66.2} & \textbf{0.3235} \\
    & \multirow{-4}{*}{1024} & \textbf{HUVR (ours)} & \underline{53.5} & \underline{65.2} & 0.3287 \\ 
    \cmidrule{2-6}
    &  & SigLIP 2 & 13.5 & 19.3 & 0.7740 \\
    &  & C-RADIOv3 & 27.3 & 37.0 & \textbf{0.5430} \\
    &  & DINOv3 & \underline{29.4} & \underline{39.2} & 0.6685 \\  
    \multirow{-8}{*}{ViT-L} & \multirow{-4}{*}{32} & \textbf{HUVR (ours)} & \textbf{30.9} & \textbf{41.1} & \underline{0.5726} \\ 
    \bottomrule
    \end{tabular}
    }
    \label{tab:dense_results}
\end{table}

For fair comparison, since these tasks tend to evaluate at higher resolution ($512{\times}512$ for ADE20K semantic segmentation, $480{\times}480$ for NYUv2 depth estimation), after we pre-train at $256{\times}256$, we have a second training stage.
We further train for 3 epochs (following DINOv3) on a mix of $256{\times}256$ and $512{\times}512$ resolution images, from ImageNet22k only.
We achieve strong results, in Table~\ref{tab:dense_results}, for both semantic segmentation and depth estimation, with both standard tokens and TinToks.

Recall that the reconstruction performance for compressed DINOv3 tokens is quite poor (Table~\ref{tab:tiny_token_results}).
Keep in mind that for a 32-dim TinTok, all information for both patch reconstruction (pixel-level) and segmentation (mid-level) must fit within 32 floating point values.
Compression helps the learning, but representation itself must be very information-dense.
In spite of these challenges, TinToks achieve the overall best performance for dense tasks.
We can further improve the dense task performance of the TinToks at the expense of reconstruction performance, and we investigate these trade-offs in Section~\ref{subsec:ablate_together} and Section~\ref{subsec:ablate_trades}.

\subsection{Our Formulation is Ideal for Predicting Image INRs}
\label{subsec:inr_results}

\begin{table}[t!]
	\centering
	\caption{\textbf{Hyper-network results.} We compare our INR hyper-network to prior works on ImageNette ($178{\times}178)$, LSUN ($256{\times}256$), and CelebA ($178{\times}178$), in terms of PSNR.}
	\resizebox{1.0\linewidth}{!}{
		\begin{tabular}{@{}l cccc@{}}
			\toprule
            Method & Epochs & ImageNette & LSUN & CelebA \\
            \midrule
            TransINR~\cite{chen2022transformers} & 4000/12.67/300 & 29.01 & 24.21 & 31.96 \\
            IPC~\cite{kim2022generalizable} & 4000/-/300 & 38.46 & - & 35.93 \\
            LA-IPC~\cite{lee2023localityawaregeneralizableimplicitneural} & 4000/-/300 & 46.10 & - & 50.74 \\
            ANR~\cite{zhang2024attention} & -/12.67/- & - & 28.30 & - \\
            Ours & \textbf{400}/\textbf{12}/\textbf{100} & \textbf{48.44} & \textbf{34.00} & \textbf{56.91} \\
            \bottomrule
        \end{tabular}
    }
    \label{tab:inr_hypernetwork_results}
\end{table}

We show that our novel INR hyper-network design achieves state-of-the-art results compared to prior arts in Table~\ref{tab:inr_hypernetwork_results}.
We set our hypernetwork here to have the same number of parameters or fewer, in terms of total encoder parameters, shared parameters, and predicted latents (details in the Appendix).
Since our network generally requires more time to train, given equal epochs, we generally use significantly fewer epochs that prior works (400 on ImageNette instead of 4000).
We thus achieve the best PSNR (the standard metric for these methods) with equal or shorter training time. 

We show how we achieve this in Table~\ref{tab:hypernetwork_design_ablation}.
The major driver of our good performance is our novel patch-wise design, which improves the performance at the cost of additional memory for forward computation. 
The global token, which we introduce mainly to facilitate tasks such as classification, also improves reconstruction performance. 
The compression and decoder are not necessary for reconstruction, but they are essential for good TinToks and well-behaved distillation, respectively.

\begin{table}[t!]
	\centering
	\caption{\textbf{Hyper-network design.} We demonstrate how our changes to the original TransINR hyper-network impact the reconstruction performance on ImageNette. For each row, we keep all changes from prior rows, such that the last row contains our full method.}
	\resizebox{1.0\linewidth}{!}{
		\begin{tabular}{@{}l cccc@{}}
			\toprule
            Method & \# Params & PSNR$\uparrow$ & SSIM$\uparrow$ & LPIPS$\downarrow$ \\
            \midrule
            TransINR + RoPE & 44.00M & 23.78 & 0.7041 & 0.3866 \\
            + only second layer~\cite{kim2022generalizable} & 43.21M & 27.15 & 0.8000 & 0.2424 \\
            + patchwise & 43.21M & 51.96 & 0.9974 & 0.0026 \\ 
            + global token & 43.41M & \textbf{53.36} & \textbf{0.9985} & \textbf{0.0007} \\
            + compression & 43.41M & 48.58 & 0.9962 & 0.0019 \\
            + decoder & 43.40M & 48.44 & 0.9954 & 0.0014 \\
            \bottomrule
        \end{tabular}
    }
    \label{tab:hypernetwork_design_ablation}
\end{table}

\subsection{Recognition and Reconstruction can Improve Together} 
\label{subsec:ablate_together}

\begin{figure}[t]
\begin{center}
\includegraphics[width=\linewidth]{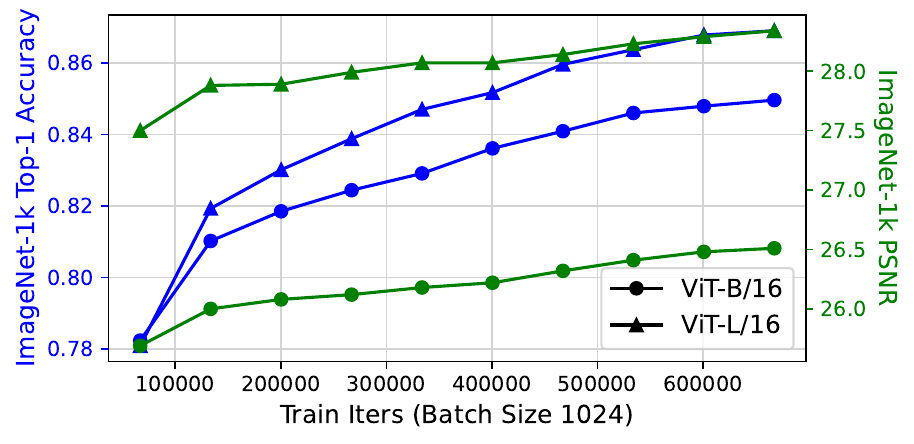} 
\caption{Training time improves both classification and reconstruction performance, although longer training yields incrementally smaller gains.}
\label{fig:training_time}
\end{center}
\end{figure}

In many ways the recognition and reconstruction performance can scale together.
One of the most obvious ways is in terms of training time, shown in Figure~\ref{fig:training_time}, although the reconstruction performance saturates more quickly than classification.
Another way is in terms of distillation teacher selection, which we explore via ablation an ablation in Table~\ref{tab:distillation_teacher_selection} where we train for 5 epochs on ImageNet22k with different teachers.
For example, the RADIOv3 ViT-L improves all 5 metrics compared to the RADIOv3 ViT-B.
In fact, for cases where the ViT-L seems worse, we find in practice that this is a result of limited training time.
Given sufficient training time (15 ImageNet22k epochs for DINOv3), there is a crossover point where distilling from the larger teacher becomes superior (see Appendix), so in our main results we distill from a ViT-L for our ViT-B, and from a ViT-H for our ViT-L.

\begin{table}[t!]
	\centering
	\caption{\textbf{Distillation teachers.} We show the effect of distilling from different teachers in terms of main token classification (d=768), tiny token classification (d=32), and reconstruction (PSNR and SSIM).}
	\resizebox{1.0\linewidth}{!}{
		\begin{tabular}{@{}llccccc@{}}
			\toprule
			\multicolumn{2}{c}{Teacher} &
			\multicolumn{2}{c}{Classification} &
			\multicolumn{1}{c}{Segmentation} &
            \multicolumn{2}{c}{Reconstruction} \\
			\cmidrule{1-2}
			\cmidrule(l){3-4}
			\cmidrule(l){5-5}
			\cmidrule(l){6-7}
            Enc. & Method & d=768 & d=32 & mIoU, d=768 & PSNR & SSIM \\
            \midrule
            & DINOv2 & 82.7 & \textbf{73.5} & 42.20 & 24.87 & 0.6950 \\
            & C-RADIOv3 & 78.3 & 47.1 & 43.32 & 25.83 & 0.7240 \\
            & SigLIP 2 & 81.3 & 58.2 & 37.87 & \textbf{26.61} & \textbf{0.7458} \\
            \multirow{-4}{*}{ViT-B} & DINOv3 & \textbf{83.2} & 69.2 & \textbf{44.01} & 25.33 & 0.7129 \\
            \midrule
            & DINOv2 & 80.9 & 74.1 & 38.37 & 25.89 & 0.7281 \\
            & C-RADIOv3 & 78.8 & 52.9 & \textbf{45.67} & \textbf{26.49} & 0.7433 \\
            & SigLIP 2 & \textbf{82.6} & 68.5 & 38.70 & 24.29 & 0.6754 \\
            \multirow{-4}{*}{ViT-L} & DINOv3 & 81.3 & \textbf{74.2} & 40.31 & 26.30 & \textbf{0.7439} \\
            \bottomrule
        \end{tabular}
    }
    \label{tab:distillation_teacher_selection}
\end{table}

\subsection{Some Design Decisions Involve Trade-offs}
\label{subsec:ablate_trades}

We can revisit Table~\ref{tab:distillation_teacher_selection} in terms of some trade-offs between performance for different tasks.
For example, the ViT-B DINOv2 teacher seems to give a better tiny token accuracy, but this is counterbalanced by a worse performance on other metrics.
RADIOv3 gives the best overall segmentation performance, but the weakest ImageNet classification, by far.
A truly optimal method would probably distill from a mixture of teachers, but we consider such engineering efforts out of scope.

\begin{table}[t!]
	\centering
	\caption{\textbf{Distillation token selection.} We show the effect of distilling for the global (cls) token only, patch tokens only, and both in terms of main token classificiation (d=768), tiny token classification (d=32), and reconstruction (PSNR and SSIM).}
	\resizebox{1.0\linewidth}{!}{
		\begin{tabular}{@{}ccccccc@{}}
			\toprule
			\multicolumn{2}{c}{Block Selection} &
			\multicolumn{2}{c}{Classification} &
			\multicolumn{1}{c}{Segmentation} &
            \multicolumn{2}{c}{Reconstruction} \\
			\cmidrule{1-2}
			\cmidrule(l){3-4}
			\cmidrule(l){5-5}
			\cmidrule(l){6-7}
            Global & Patch & d=768 & d=32 & mIoU, d=768 & PSNR & SSIM \\
            \midrule
             &  & 11.4 & 1.9 & 4.18 & \textbf{28.42} & \textbf{0.7973} \\
            \checkmark &  & 82.9 & \textbf{71.3} & 39.29 & 26.42 & 0.7468 \\
             & \checkmark & 81.9 & 65.5 & \textbf{45.23} & 26.95 & 0.7580 \\
            \checkmark & \checkmark & \textbf{83.2} & 69.3 & 43.68 & 25.33 & 0.7133 \\
            \bottomrule
        \end{tabular}
    }
    \label{tab:distillation_token_selection}
\end{table}

In Table~\ref{tab:distillation_token_selection} we compare our chosen distillation approach (where we compute distillation losses to both the global token and patch tokens) to alternatives where we distill only to the global token, only to patch tokens, or neither.
For settings, we use a DINOv3 ViT-B teacher and train for 5 ImageNet22k epochs only.
In general, the impact of the distillation target tokens on patch performance makes sense.
Distillation to the global token is helpful for classification, and distillation to patches is helpful for segmentation.
Interestingly, the overall $d=768$ classification is best when we distill to all tokens.
However, the reconstruction suffers substantially.
We ultimately choose this setting anyway, and counterbalance the falling reconstruction metrics with longer training and distillation from a larger teacher model.

\begin{table}[t!]
	\centering
	\caption{\textbf{Distillation block selection.} We ablate selection of which block output we choose for the distillation. We try 2 blocks each for the Encoder and the Decoder and report performance of the standard token (d=768) and tiny token (d=32) for ImageNet linear probing classification, and of the tiny token for reconstruction (PSNR, SSIM). Recall that the encoder and decoder have 12 and 4 blocks, respectively.}
	\resizebox{1.0\linewidth}{!}{
		\begin{tabular}{@{}cccccc@{}}
			\toprule
			\multicolumn{2}{c}{Block Selection} &
			\multicolumn{2}{c}{Classification} &
            \multicolumn{2}{c}{Reconstruction} \\
			\cmidrule{1-2}
			\cmidrule(l){3-4}
			\cmidrule(l){5-6}
            Encoder & Decoder & d=768 & d=32 & PSNR & SSIM \\
            \midrule
            11 & 1 & 82.5 & 69.7 & 26.06 & 0.7370 \\
            11 & 4 & 82.9 & \textbf{72.6} & 25.40 & 0.7143 \\
            12 & 1 & \textbf{83.2} & 68.8 & \textbf{26.18} & \textbf{0.7454} \\
            12 & 4 & 83.1 & 69.4 & 25.36 & 0.7140 \\
            \bottomrule
        \end{tabular}
    }
    \label{tab:distillation_block_selection}
\end{table}

We can select any combination of blocks for distillation, and we perform a limited investigation in Table~\ref{tab:distillation_block_selection}, where we train for 5 ImageNet22k epochs, using a DINOv3 ViT-B teacher.
The best setting for reconstruction and standard classification (12, 1) is the worst for tiny token classification.
Distilling to (11, 4) leads to better TinTok classification, and (12, 1) yields the best reconstruction. 
However, we ultimately opt to distill to the last block of each network (12, 4), both as a good middle ground and for the sake of simplicity.

\begin{table}[t!]
	\centering
	\caption{\textbf{Distillation design.} We show the impact of increasing the tiny token size to match the original tokens, chaning the INR hidden dimension from 256 to 512, changing the number of INR layers from 4 to 3 or 5, reducing from 4 transformer decoder layers to 1, or removing the attention from the transformer decoder.}
	\resizebox{1.0\linewidth}{!}{
		\begin{tabular}{@{}lccccc@{}}
			\toprule
            &
			\multicolumn{2}{c}{Classification} &
			\multicolumn{1}{c}{Segmentation} &
            \multicolumn{2}{c}{Reconstruction} \\
			\cmidrule{1-2}
			\cmidrule{2-3}
			\cmidrule(l){4-4}
			\cmidrule(l){5-6}
            Setting & d=768 & d=32 & mIoU, d=768 & PSNR & SSIM \\
            \midrule
            Default & \textbf{83.1} & 69.7 & 43.85 & 25.35 & 0.7135 \\
            Tiny dim=768 & \textbf{83.1} & 69.1 & 44.00 & \textbf{25.65} & \textbf{0.7243} \\
            INR w/ dim=512 & 82.3 & 70.8 & 23.36 & 25.30 & 0.7119 \\
            INR w/ 3 layers & \textbf{83.1} & 69.6 & 43.36 & 25.08 & 0.7015 \\
            INR w/ 5 layers & \textbf{83.1} & 70.2 & \textbf{44.44} & 25.48 & 0.7188 \\
            Trans. w/ 1 layer & \textbf{83.1} & 69.6 & 43.65 & 24.78 & 0.6962 \\
            Trans. w/o att. & \textbf{83.1} & \textbf{73.5} & 43.83 & 24.90 & 0.6943 \\
            \bottomrule
        \end{tabular}
    }
    \label{tab:decoder_design_ablation}
\end{table}

We finally examine certain decoder design choices in Table~\ref{tab:decoder_design_ablation}.
We show how some choices that would increase computational cost, like increasing the INR hidden dimension from 256 to 512, actually hurt results.
Increasing the token dimension in the decoder helps the reconstruction metrics, but at the expense of both speed and TinTok classification.
The most significant decision involves removing attention from the hyper-network decoder.
This converts the decoder from a transformer, to a multi-layer perceptron with residual connections.
This improves the TinTok classification, without affecting the standard-sized token classification or segmentation.
However, it has a significant negative impact on the reconstruction performance.
In spite of this, for our TinTok experiments (such as in Table~\ref{tab:tiny_token_results}), we opt to use this setting, and we mainly to counter-balance the negative effect with more training time.

\subsection{Limitations}

Our pre-training does not operate at the same scale or scope as prior image representation methods.
Compared to SigLIP 2, we train with less data for less time.
Compared to DINOv3, we do not train with specialized curated data.
Compared to Radio, we focus more on the unified modeling aspect, and do not engineer distillation from multiple models.
Due to this, while HUVR and TinyTokens have the best performance for many benchmarks, and are the only method with native reconstruction capability, there are some datasets and metrics for which other methods are superior.
Potential application to tasks with Vision Language Models (VLMs) would require text-aligned pre-training, and could be future work.
\section{Conclusion}
\label{sec:conclusion}

We propose HUVR, an INR hyper-network for unified universal image representation. 
Not only does HUVR compare favorably with prior works for unsupervised learning for image recognition, it also yields compressed TinToks which support recognition and reconstruction.
We demonstrate the utility of the method for various datasets and embeddings sizes for tasks including classification, segmentation, and generation.
We hope that this work enables further exploration in the areas of unified representation learning, compressed representations, and implicit neural representation as a viable strategy for universal vision encoding.

\clearpage

\maketitlesupplementary

\section{Implicit Neural Representation and Hyper-network Tutorial}

\subsection{Image INR Basics}

Recall from Section~\ref{subsec:hyper-networks} that an INR is a function $f_\theta:X\rightarrow I$ for coordinates $X$ and signal (image) $I$.
In practice, this INR might be a multi-layer perceptron, with $n$ feedforward layers separated by some non-linearity, such as a ReLU.
In addition, INRs perform best when they do not operate directly on coordinate inputs.
Instead, the coordinate tuples ($x$, $y$) are first encoded with some sinusoidal or Fourier transformation~\cite{sitzmann2020implicit,tancik2020fourfeat}.
So, for some image $I$ with width $W=1024$ and $H=1024$, we might represent the image with an MLP with $n=5$ layers, with positional encoding dimension $d_\text{pos}=128$, hidden dimension $d_\text{hid}=256$, and output dimension $d_\text{output}=3$ (the number of color channels).

When training an INR, we use a paradigm referred to as ``internal learning.''
Unlike typical machine learning, there is no distinction between training and testing data, and instead the goal is to overfit $f_\theta$ to $I$ during training, such that $f_\theta$ ``memorizes'' $I$ and can perfectly reconstruct it during inference.
This is why it is called ``implicit'' representation -- $I$ is totally stored in $\theta$.
To train such a network, we perform stochastic gradient descent with all pixel coordinates $X$ as inputs and all pixel color values $I$ as outputs.
After iterating over all pixels multiple times (sometimes hundreds of times, depending on the complexity of the signal), an INR of sufficient size can achieve visually lossless reconstruction.

Sometimes these networks are trained on 3D scenes (with many images from different viewpoints), and in such cases, they are expected to perform novel view synthesis (generate an image from an unseen viewpoint).
In the image-specific regime, where each network represents a single image (and is trained on that image, only), INRs can also perform some ``generative'' tasks.
These include super-resolution, where we can provide interpolated coordinates to the network as input, and it will give the corresponding color prediction as output.
Following the same approach, we can give coordinates outside the training range, and the network will perform outpainting.

\subsection{Hyper-networks}

As we mention previously, the main problem with these networks is they require extensive training time.
While some approaches try to accelerate the training, hyper-networks try avoid the per-network training entirely.
Instead, hyper-networks rely on a pre-training stage to train a network $h:I\rightarrow f_{\theta'}$.
While we already explain this to some extent in Section~\ref{subsec:hyper-networks}, we provide even more details here.

The most counterintuitive aspect of hyper-networks is that they do not actually predict the entire implicit neural network at inference time.
Instead, the actual INR is defined at the same time as the hyper-network, and the weights are learned during the hyper-network pre-training.
However, since this INR is fit on the entire dataset, it actually does not represent an image.
If one were to perform inference directly with that INR, they would get an image with noise resembling a sort of dataset ``average image.''

Considering that we train the ``base'' INR weights, $\theta_b$, on the entire dataset, the role of $h$ is obviously not to predict the entire INR weights, $\theta'$.
Instead, $h$ simply needs to use input $I$ to modulate the weights of $\theta_b$ such that inference for $X$ with the resulting $\theta'$ yields the actual input image $I$. 
In the original Trans-INR~\cite{chen2022transformers}, $h$ predicts a unique matrix to modify each layer of $\theta_b$.
However, IPC~\cite{kim2022generalizable} shows that it is sufficient to modulate only the second layer, and since we adopt that practice in this paper, we will explain the process from the perspective that we only modulate a single layer.
In other words, if we define an INR with $n=5$, we only modulate $W_2$, and $\theta'$ uses $W_1$, $W_3$, $W_4$, and $W_5$ directly from $\theta_b$.
The modulation for $\theta'$ proceeds as we explain in Section~\ref{subsec:hyper-networks}, with the second layer given by $W_2 \odot M_2$.

\section{Ablations}

\begin{table*}[t!]
	\centering
	\caption{\textbf{Distillation Teacher Size Crossover Effect}. When distilling from DINOv3, training with a teacher larger than the student initially yields poorer results than training with a similar-size teacher. However, after training for enough time, training with the larger teacher is superior. For each metric, we highlight the value for the earliest epoch where the ViT-L teacher gives a better result than the ViT-B at the same epoch. We give TinTok results ($d=32$) for ImageNet only, but mostly focus on standard tokens.
    }
	\resizebox{1.0\linewidth}{!}{
		\begin{tabular}{@{}lc ccccccccc cc cc@{}}
			\toprule
			\multicolumn{2}{c}{Settings} &
			\multicolumn{9}{c}{Classification} &
			\multicolumn{2}{c}{Segment. (ADE20K)} &
			\multicolumn{2}{c}{Recon. (INet)} \\
			\cmidrule{1-2}
			\cmidrule(l){3-11}
			\cmidrule(l){12-13}
			\cmidrule(l){14-15}
    Teacher & Epochs & INet ($d=32$) & INet & ReaL & ONet & Cars & CUB & DTD & Flowers & Food & mIoU & mAcc & PSNR & SSIM \\
    \midrule    
    \cellcolor{white} & 10 & 70.6 & 82.5 & 87.7 & 53.3 & 92.8 & 88.3 & 82.6 & 99.7 & 91.9 & 45.74 & 58.28 & 25.08 & 0.7001 \\
    \cellcolor{white} & 20 & 71.4 & 83.2 & 88.2 & 54.6 & 92.9 & 88.3 & 83.6 & 99.6 & 92.4 & 46.11 & 58.52 & 25.18 & 0.7027 \\
    \cellcolor{white} & 30 & 71.6 & 83.6 & 88.3 & 56.0 & 93.1 & 88.9 & 84.1 & 99.6 & 92.7 & 46.85 & 59.44 & 25.22 & 0.7036 \\
    \cellcolor{white} & 40 & 72.5 & 84.0 & 88.5 & 57.8 & 93.3 & \textbf{89.3} & 84.3 & 99.7 & 93.2 & 46.87 & 58.69 & 25.43 & 0.7086 \\
    \multirow{-5}{*}{ViT-B} \cellcolor{white} & 50 & 72.8 & 84.1 & 88.6 & 58.3 & \textbf{93.4} & \textbf{89.3} & \textbf{84.9} & 99.6 & 93.2 & 47.17 & 58.72 & 25.50 & 0.7108 \\
    \midrule
    \cellcolor{white} & 10 & 70.6 & 80.3 & 86.2 & 47.3 & 91.1 & 87.3 & 80.8 & 99.5 & 90.4 & 44.82 & 56.74 & \cellcolor{green} 25.65 & \cellcolor{green} 0.7173 \\
    \cellcolor{white} & 20 & \cellcolor{green} 72.8 & 82.1 & 87.4 & 51.4 & 91.9 & 88.1 & 82.6 & 99.6 & 91.6 & \cellcolor{green} 46.33 & \cellcolor{green} 59.06 & 25.70 & 0.7194 \\
    \cellcolor{white} & 30 & 74.5 & 83.1 & 88.0 & 55.2 & 92.4 & 88.9 & 82.8 & 99.7 & 92.7 & 47.35 & 58.86 & 25.68 & 0.7189 \\
    \cellcolor{white} & 40 & 75.9 & \cellcolor{green} 84.1 & \cellcolor{green} 88.7 & \cellcolor{green} 58.9 & 92.9 & 89.2 & \cellcolor{green} 84.8 & \cellcolor{green} \textbf{99.8} & \cellcolor{green} 93.7 & 48.41 & 60.04 & 25.95 & 0.7250 \\
    \multirow{-5}{*}{ViT-L} \cellcolor{white} & 50 & \textbf{76.6} & \textbf{84.6} & \textbf{88.9} & \textbf{60.8} & 93.1 & \cellcolor{green} \textbf{89.3} & 84.7 & 99.7 & \textbf{94.2} & \textbf{48.71} & \textbf{60.51} & \textbf{26.02} & 0.7275 \\
    \bottomrule
    \end{tabular}
    }
    \label{tab:distillation_size_crossover}
\end{table*}

\subsection{Distillation Teacher Size Crossover Point}

When we train the model with a larger teacher, it requires more epochs before it outperforms the smaller teacher.
However, given sufficiently long training time, larger teachers are superior.
We illustrate this crossover effect for our ViT-B/16 in Table~\ref{tab:distillation_size_crossover}.
We use the attention-free decoder for this experiment (hence the numbers do not exactly match with tables in the main paper).
Note that the crossover effect impacts standard-size tokens ($d=768$) more than TinToks ($d=32$).
The reconstruction and classification of the TinToks is better using large teachers, regardless of the number of training epochs.
However, the standard tokens require many training epochs to justify using the DINOv3 ViT-L teacher instead of the DINOv3 ViT-B teacher (up to 40 epochs for classification, 10 epochs for segmentation).

\begin{table}[t!]
	\centering
	\caption{\textbf{Distillation loss weights.} We show the effect of different distillation weights in terms of main token classification (d=768) and tiny token classification (d=32) on ImageNet, main token segmentation (d=768) on ADE20K, and reconstruction (PSNR) on ImageNet. We first show our main setting, then demonstrate the impact of independently increasing the decoder weights ($\alpha_\text{dec})$, the global token weights ($\alpha_g)$, and the patch token weights ($\alpha_p$).}
	\resizebox{1.0\linewidth}{!}{
		\begin{tabular}{@{}ccccccccc@{}}
			\toprule
			\multicolumn{4}{c}{Loss Weights} &
			\multicolumn{2}{c}{Classification} &
			\multicolumn{1}{c}{Segmentation} &
            \multicolumn{1}{c}{Recon.} \\
			\cmidrule{1-4}
			\cmidrule(l){5-6}
			\cmidrule(l){7-7}
			\cmidrule(l){8-8}
            $\alpha_{g, \text{enc}}$ & $\alpha_{p, \text{enc}}$ & $\alpha_{g, \text{dec}}$ & $\alpha_{p, \text{dec}}$ & d=768 & d=32 & mIoU (d=768) & PSNR \\
            \midrule
            2.0 & 2.0 & 0.5 & 0.5 & 83.0 & 68.2 & 43.68 & 25.84 \\
            \cellcolor{lightgray} 1.0 & \cellcolor{lightgray} 1.0 & \cellcolor{lightgray} 0.25 & \cellcolor{lightgray} 0.25 & 82.8 & \textbf{69.2} & 43.97 & \textbf{26.75} \\
            2.0 & 2.0 & \cellcolor{lightgray} 1.0 & \cellcolor{lightgray} 1.0 & 83.0 & 68.7 & 43.42 & 25.19 \\
            \cellcolor{lightgray} 4.0 & 2.0 & \cellcolor{lightgray} 1.0 & 0.5 & 82.9 & 68.2 & 42.83 & 25.17 \\
            2.0 & \cellcolor{lightgray} 4.0 & 0.5 & \cellcolor{lightgray} 1.0 & \textbf{83.2} & 67.9 & \textbf{44.97} & 25.49 \\
            \bottomrule
        \end{tabular}
    }
    \label{tab:distillation_loss_weights}
\end{table}

\subsection{Distillation Loss Weights}

We perform an ablation for different distillation loss weights (Equation~\ref{eq:distillation}) in Table~\ref{tab:distillation_loss_weights}.
For this experiment, we train for only 5 ImageNet22k epochs.
Evaluations follow the same linear probing and reconstruction settings as in our other experiments.

\begin{table}[t!]
	\centering
	\caption{\textbf{Reconstruction objective.} We train our system with no reconstruction loss (no INR prediction, and no image prediction), compared to our standard system, for 5 epochs on ImageNet22k. The network without INR prediction and associated loss is fundamentally incapable of reconstruction.}
	\resizebox{0.5\linewidth}{!}{
		\begin{tabular}{@{}lccc@{}}
			\toprule
            &
			\multicolumn{2}{c}{Classification} &
            \multicolumn{1}{c}{Recon.} \\
			\cmidrule(l){2-3}
			\cmidrule(l){4-4}
            INR & d=768 & d=32 & PSNR \\
            \midrule
            $\times$ & \textbf{83.3} & 64.2 & - \\
            $\checkmark$ & 83.0 & \textbf{68.2} & \textbf{25.84} \\
            \bottomrule
        \end{tabular}
    }
    \label{tab:inr_vs_no_inr}
\end{table}

\subsection{Hyper-network Helps Classification}

We can train HUVR without the loss for INR prediction and image reconstruction.
The resulting model cannot reconstruct images, so its embeddings and TinToks cannot be used for image compression or generation.
It does not learn a unified representation.
Additionally, as we show in Table~\ref{tab:inr_vs_no_inr}, the TinToks classification is significantly worse without the image reconstruction objective.
This shows that the hyper-network prediction and image reconstruction are synergistic with the recognition task.

\section{Detailed Settings}

\subsection{Main Experiments}

\begin{table}[t!]
	\centering
	\caption{\textbf{Encoder architectures.} We show settings for our ViT-B/16 and ViT-L/16.}
	\resizebox{0.9\linewidth}{!}{
		\begin{tabular}{@{}lcc@{}}
			\toprule
			Backbone & ViT-B/16 & ViT-L/16 \\
            \midrule
            \# Parameters & $85.9$M & $304$M \\
            \# Blocks & 12 & 24 \\
            Position Embedding & RoPE & RoPE \\
            Register Tokens & None & None \\
            Token Dimension & 768 & 1024 \\
            Feedforward Dimension & 2048 & 3072 \\
            Feedforward Type & SwiGLU & SwiGLU~\cite{shazeer2020gluvariantsimprovetransformer} \\
            Attention Heads & 16 & 16 \\
            Attention Head Dimension & 48 & 64 \\
            \bottomrule
        \end{tabular}
    }
    \label{tab:encoder_settings}
\end{table}

\begin{table}[t!]
	\centering
	\caption{\textbf{Decoder architectures.} We show settings for sample \textbf{decoders} we can use with our ViT-B and ViT-L. For some ablations, we instead use token dimension 512 for the sake of speed.}
	\resizebox{0.9\linewidth}{!}{
		\begin{tabular}{@{}lcc@{}}
			\toprule
			Backbone & ViT-B/16 & ViT-L/16 \\
            \midrule
            \# Parameters & $50.3$M & $68.2$M \\
            \# Blocks & 4 & 4 \\
            Position Embedding & Sinusoidal & Sinusoidal \\
            Token Dimension & 1024 & 1280 \\
            Feedforward Dimension & 4096 & 4096 \\
            Attention Heads & 16 & 16 \\
            Attention Head Dimension & 64 & 80 \\
            \bottomrule
        \end{tabular}
    }
    \label{tab:decoder_settings}
\end{table}

\begin{table}[t!]
	\centering
	\caption{\textbf{INR architectures.} We show settings for the base patch-wise INR $f_{\theta_b}$.}
	\resizebox{0.75\linewidth}{!}{
		\begin{tabular}{@{}lc@{}}
			\toprule
			\multicolumn{2}{c}{Settings} \\
            \midrule
            \# Parameters & $275$k \\
            Position Embedding & Sinusoidal \\
            Embedding Dimension & 128 \\
            \# MLP Layers & 3 \\
            MLP Dimension & 256 \\
            \# Conv layers & 1 \\
            Kernel Size & 3 \\
            Padding & 1 \\
            Stride & 1 \\
            Upsample & PixelShuffle~\cite{shi2016realtimesingleimagevideo} \\
            Activation & ReLU \\
            \bottomrule
        \end{tabular}
    }
    \label{tab:inr_settings}
\end{table}

Our ViT-B and ViT-L have $85.9$M and $304$M parameters, respectively. 
We show full settings for our ViT-B and ViT-L in Table~\ref{tab:encoder_settings}.
In places where we change the design compared to prior works, we make sure to keep total parameters equivalent (for example, our ViT-B has fewer than 86M parameters).
Our decoder parameters and designs vary depending on the experiment settings, but we show decoder settings for our main experiments in Table~\ref{tab:decoder_settings}.
Note that these are ``worst case'' sizes, where we exchange the increased parameter count for marginal improvement to reconstruction performance (see Table~\ref{tab:decoder_design_ablation}. Reducing from 4 layers to 1 ($4\times$ fewer parameters) does not incur a penalty for classification or segmentation. 
Similarly, recall that even with a single layer decoder, we can have good reconstruction performance~\ref{tab:inr_hypernetwork_results}.
Therefore, we hypothesize that future work that focuses on efficiency can reduce the parameter count and improve both training and inference efficiency.

We also give settings for our INR with learnable weights $\theta_b$ in Table~\ref{tab:inr_settings}.
Note that we mention a convolutional layer and PixelShuffle.
This is because we do not use a pure MLP INR, nor do we take in every coordinate at inference.
Instead, following intuitions from prior INR works~\cite{chen2021nerv,maiya2023nirvana}, we use strided coordinates ($\text{stride}=4$) as inputs to $f_{\theta'}$ to reduce the computation by $16{\times}$.
We use the learnable convolutional layer with PixelShuffle to upsample from $(H/4){\times}(W/4)$ features to the $H{\times}W$ image.

We use a base learning rate ($\text{lr}$) of $0.0005$ which we rescale by multiplying $\text{lr} = \text{lr}_\text{base} * B / 256$ for some batch size, $B$.
In our standard setting we train for 50 epochs (or the equivalent number of iterations, based on 13.7M training images in ImageNet22k).
We use the AdamW~\cite{loshchilov2019decoupledweightdecayregularization} optimizer, a cosine annealing learning rate after linear warm-up for 5 epochs, and we clip gradients with a norm of $0.01$.
For data augmentations, we take a random resized crop, with a minimum rescale ratio of $0.2$.
We then resize to our input resolution, $256{\times}256$ for pretraining.
For our model, we use normalization with means $(0.5, 0.5, 0.5)$ and standard deviations $(0.5, 0.5, 0.5)$.
When interacting with other models, either as teachers or when performing evaluations with them, we use their original image normalization settings.

Aside from our encoder designing and training, we also have some other settings for the components of our pipeline that predict the INR, and for the INR itself.
First, we have a downsample layer that consists of a LayerNorm~\cite{ba2016layernormalization} followed by fully-connected layer which projects from $d_\text{ViT}$ to $d_t$.
Our upsample layer is the same but projects from $d_t$ to $d_\text{dec}$ where $d_\text{dec}$ is the token dimension of the transformer decoder that we use to predict matrices to modulate the INR.
We also need learnable modules to project from $d_\text{dec}$ to $d_\text{in}$ and these also have learnable LayerNorms.
We learn two such layers, one for the global token, and another for the patch tokens.
We also have layers to project between our tokens and the teacher representation space, which also feature a single LayerNorm followed by a linear layer to project from $d_\text{ViT}$ or $d_\text{dec}$ to $d_\text{teacher}$.
For our ViT-B hyper-network, the ViT encoder itself has $85.9$M parameters, the learnable $\theta_b$ has $275$k parameters, and the remaining components (distillation projection layers, upsample/downsample, projection from decoder to INR) have $3.16$M learnable parameters (of which only $305$k are used at inference).

\subsection{INR Hyper-network Comparisons}

When we indicate the training epochs in Table~\ref{tab:inr_hypernetwork_results}, we give the epochs for training on ImageNette/LSUN/CelebA, respectively.
For fractional epochs, those papers train for a set number of iterations rather than epochs, and we convert to epochs for comparison.
We only train using whole-valued epochs, so we train for 12 instead of 12.67 epochs on LSUN Churches.

The numbers in Table~\ref{tab:inr_hypernetwork_results} come from multiple sources.
For TransINR, we reproduce the results using the model settings from the original paper~\cite{chen2022transformers} and the indicated number of epochs, AdamW optimizer, batch size of $16$, and learning rate of $0.0001$.
For IPC and LA-IPC we use the numbers from LA-IPC~\cite{lee2023localityawaregeneralizableimplicitneural}.
For INR, we use the number from ANR~\cite{zhang2024attention}.
For ours, we use the indicated number of epochs and the same optimizer, batch size, and learning rate as with our baseline TransINR reproduction.

We make the settings fair.
With hypernetworks, there are several factors to consider, mainly (i) the size of the encoder $E$, (ii) the size of the learnable base INR weights $\theta_b$, and (iii) the number of ``unique parameters'', which is the smallest size of the portion of the weight tokens that is necessary to modulate $\theta_b$ to produce $\theta'$.
The explain the latter with an example, consider an encoder $E$ that predicts $100$ tokens with $d=768$, but only $50$ are used to modulate the INR weights.
In this case, the size of the unique parameters would be $38,400 = 50 * 768$.
However, note that we said the ``smallest size.''
If the encoder, $E$, predicts the tokens at $d=768$, but these are downsampled to a smaller dimension, such as $d=32$, before the modulation, then we would say the size of the unique parameters is $1600 = 50 * 32$.

\noindent\textbf{Transformer Encoder Parameters.} 
The prior works use the same design for the transformer encoder, and we match this design with one exception.
Since our model has a decoder, we must reduce the size of the encoder to maintain fairness.
We accomplish this by removing a single layer from $E$, and our decoder for the hyper-network comparisons consists of a single layer with the exact same settings as $E$ (aside from the number of layers, which is 1).
So, the encoders are all the same size, except for when we downsample for TinTok compression.
In these cases, our model adds an extra 393k parameters for the learnable downsample from $d=768$ to $d=256$ and learnable upsample from $d=256$ to $d=768$.

\noindent\textbf{Base INR Parameters.} 
Our $\theta_b$ consists of 230k learnable parameters, for all 3 datasets.
We use this same exact design for our TransINR reproduction, and IPC and LA-IPC use these same settings for ImageNette and CelebA.
ANR's learnable INR consists of 2757k parameters for LSUN.
Our patch-wise INR is significantly smaller than the image-wise INRs, even though these ultimately represent the exact same data.
However, our patch-wise INR formulation can result in less efficiency during forward computation, since our modulation procedure will yield $p$ unique INRs, where $p$ is the number of image patches ($256$ for LSUN, $400$ for ImageNette/CelebA).
This has slightly higher time costs and a larger memory footprint, but due to our TinTok compression, our storage size is smaller (see Unique Parameters, next paragraph).

\noindent\textbf{Unique Parameters.} 
ANR predicts 131k unique parameters to modulate their INRs for LSUN, while ours, IPC, and LA-IPC predict 65.5k (half as many).
TransINR predicts 58.1k unique parameters, but uses portions of these to modulate each layer separately, compared to ours, IPC, and LA-IPC which modulate only the second layer.

\section{Diffusion with HUVR}

\begin{figure}[t]
\begin{center}
\includegraphics[width=\linewidth]{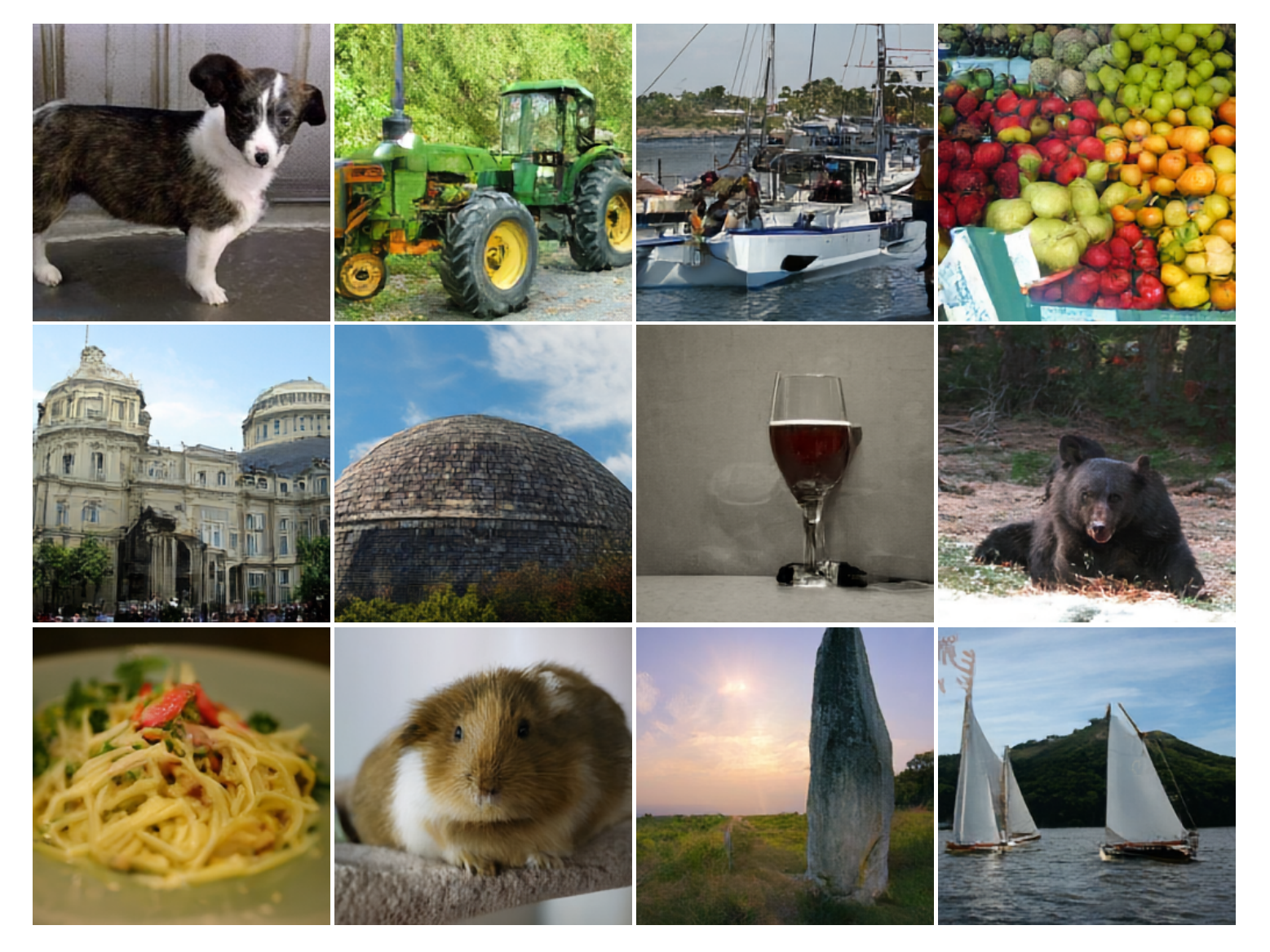}
\caption{\textbf{Generated samples with HUVR embeddings.} We use a DiT-XL train on HUVR embeddings with TinTok $d_t=256$. This HUVR is trained with LPIPS and SSIM losses in addition to the pixel-wise and DINOv3 MSE losses. Compared to our Table~\ref{tab:diffusion_results}, this DiT is trained for 4500k steps instead of 400k steps. For reference, DiT-XL/2 trains their final model for 7000k steps. These results demonstrate many degradations and artifacts, but we hope they work as a proof-of-concept to convey the promise of HUVR for generation. Future work could apply techniques, such as those in RAE~\cite{zheng2025diffusiontransformersrepresentationautoencoders}, to improve the quality.}
\label{fig:diffusion_results}
\end{center}
\end{figure}

We show visual diffusion examples in Figure~\ref{fig:diffusion_results}.
These initial results, combined with our comparisons to the VAE reconstruction (Table~\ref{tab:tiny_token_results}) and VAE-based DiT generation (Table~\ref{tab:diffusion_results}), are meant to show the potential of our method for generation.
However, we acknowledge these results are significantly worse than the current generative state-of-the-arts.

In most respects, this is because we hypothesize we have optimized neither the diffusion model nor the diffusion process for our architecture.
For example, we have both patch tokens and a global token, and the global token is responsible for combining with the patch token to modulate every single patch INR.
In spite of this, our implementation of DiT handles the global and patch tokens equivalently.
Aside from this, there are many other strategies that could improve performance given our larger token size (we use $d_t=256$ here), larger patch size (16 for ours instead of 8 for VAE), and the semantic knowledge in our tokens.
We leave these for future work, and hope our results here are sufficient to demonstrate this as a compelling area for future research.




\begin{thebibliography}{120}
\providecommand{\natexlab}[1]{#1}
\providecommand{\url}[1]{\texttt{#1}}
\expandafter\ifx\csname urlstyle\endcsname\relax
  \providecommand{\doi}[1]{doi: #1}\else
  \providecommand{\doi}{doi: \begingroup \urlstyle{rm}\Url}\fi

\bibitem[Assran et~al.(2022)Assran, Caron, Misra, Bojanowski, Bordes, Vincent, Joulin, Rabbat, and Ballas]{assran2022masked}
Mahmoud Assran, Mathilde Caron, Ishan Misra, Piotr Bojanowski, Florian Bordes, Pascal Vincent, Armand Joulin, Michael Rabbat, and Nicolas Ballas.
\newblock Masked siamese networks for label-efficient learning, 2022.

\bibitem[Ba et~al.(2016)Ba, Kiros, and Hinton]{ba2016layernormalization}
Jimmy~Lei Ba, Jamie~Ryan Kiros, and Geoffrey~E. Hinton.
\newblock Layer normalization, 2016.

\bibitem[Bai et~al.(2023)Bai, Bai, Chu, Cui, Dang, Deng, Fan, Ge, Han, Huang, et~al.]{bai2023qwen}
Jinze Bai, Shuai Bai, Yunfei Chu, Zeyu Cui, Kai Dang, Xiaodong Deng, Yang Fan, Wenbin Ge, Yu Han, Fei Huang, et~al.
\newblock Qwen technical report.
\newblock \emph{arXiv preprint arXiv:2309.16609}, 2023.

\bibitem[Bao et~al.(2022)Bao, Dong, Piao, and Wei]{bao2022beit}
Hangbo Bao, Li Dong, Songhao Piao, and Furu Wei.
\newblock Beit: Bert pre-training of image transformers, 2022.

\bibitem[Barbu et~al.(2019)Barbu, Mayo, Alverio, Luo, Wang, Gutfreund, Tenenbaum, and Katz]{barbu2019objectnet}
Andrei Barbu, David Mayo, Julian Alverio, William Luo, Christopher Wang, Dan Gutfreund, Josh Tenenbaum, and Boris Katz.
\newblock Objectnet: A large-scale bias-controlled dataset for pushing the limits of object recognition models.
\newblock \emph{Advances in neural information processing systems}, 32, 2019.

\bibitem[Bardes et~al.(2021)Bardes, Ponce, and LeCun]{Bardes2021VICRegVR}
Adrien Bardes, Jean Ponce, and Yann LeCun.
\newblock Vicreg: Variance-invariance-covariance regularization for self-supervised learning.
\newblock \emph{ArXiv}, abs/2105.04906, 2021.

\bibitem[Beyer et~al.(2020)Beyer, Hénaff, Kolesnikov, Zhai, and van~den Oord]{beyer2020imagenet}
Lucas Beyer, Olivier~J. Hénaff, Alexander Kolesnikov, Xiaohua Zhai, and Aäron van~den Oord.
\newblock Are we done with imagenet?, 2020.

\bibitem[Bossard et~al.(2014)Bossard, Guillaumin, and Van~Gool]{bossard14}
Lukas Bossard, Matthieu Guillaumin, and Luc Van~Gool.
\newblock Food-101 -- mining discriminative components with random forests.
\newblock In \emph{European Conference on Computer Vision}, 2014.

\bibitem[Caron et~al.(2018)Caron, Bojanowski, Joulin, and Douze]{caron2018deep}
Mathilde Caron, Piotr Bojanowski, Armand Joulin, and Matthijs Douze.
\newblock Deep clustering for unsupervised learning of visual features.
\newblock In \emph{Proceedings of the European Conference on Computer Vision (ECCV)}, pages 132--149, 2018.

\bibitem[Caron et~al.(2020)Caron, Misra, Mairal, Goyal, Bojanowski, and Joulin]{caron2020unsupervised}
Mathilde Caron, Ishan Misra, Julien Mairal, Priya Goyal, Piotr Bojanowski, and Armand Joulin.
\newblock Unsupervised learning of visual features by contrasting cluster assignments.
\newblock \emph{Advances in Neural Information Processing Systems}, 33:\penalty0 9912--9924, 2020.

\bibitem[Caron et~al.(2021)Caron, Touvron, Misra, J\'egou, Mairal, Bojanowski, and Joulin]{caron2021emerging}
Mathilde Caron, Hugo Touvron, Ishan Misra, Herv\'e J\'egou, Julien Mairal, Piotr Bojanowski, and Armand Joulin.
\newblock Emerging properties in self-supervised vision transformers.
\newblock In \emph{Proceedings of the International Conference on Computer Vision (ICCV)}, 2021.

\bibitem[Chen et~al.(2021)Chen, He, Wang, Ren, Lim, and Shrivastava]{chen2021nerv}
Hao Chen, Bo He, Hanyu Wang, Yixuan Ren, Ser~Nam Lim, and Abhinav Shrivastava.
\newblock Nerv: Neural representations for videos.
\newblock \emph{Advances in Neural Information Processing Systems}, 34:\penalty0 21557--21568, 2021.

\bibitem[Chen et~al.(2022)Chen, Gwilliam, He, Lim, and Shrivastava]{chen2022cnerv}
Hao Chen, Matt Gwilliam, Bo He, Ser-Nam Lim, and Abhinav Shrivastava.
\newblock Cnerv: Content-adaptive neural representation for visual data, 2022.

\bibitem[Chen et~al.(2023)Chen, Gwilliam, Lim, and Shrivastava]{chen2023hnerv}
Hao Chen, Matthew Gwilliam, Ser-Nam Lim, and Abhinav Shrivastava.
\newblock Hnerv: A hybrid neural representation for videos.
\newblock In \emph{Proceedings of the IEEE/CVF Conference on Computer Vision and Pattern Recognition}, pages 10270--10279, 2023.

\bibitem[Chen et~al.(2024)Chen, Xie, Lim, and Shrivastava]{chen2024fastencodingdecodingimplicit}
Hao Chen, Saining Xie, Ser-Nam Lim, and Abhinav Shrivastava.
\newblock Fast encoding and decoding for implicit video representation, 2024.

\bibitem[Chen et~al.(2020{\natexlab{a}})Chen, Kornblith, Norouzi, and Hinton]{chen2020simple}
Ting Chen, Simon Kornblith, Mohammad Norouzi, and Geoffrey Hinton.
\newblock A simple framework for contrastive learning of visual representations.
\newblock In \emph{International conference on machine learning}, pages 1597--1607. PMLR, 2020{\natexlab{a}}.

\bibitem[Chen et~al.(2020{\natexlab{b}})Chen, Kornblith, Swersky, Norouzi, and Hinton]{chen2020big}
Ting Chen, Simon Kornblith, Kevin Swersky, Mohammad Norouzi, and Geoffrey~E Hinton.
\newblock Big self-supervised models are strong semi-supervised learners.
\newblock \emph{Advances in neural information processing systems}, 33:\penalty0 22243--22255, 2020{\natexlab{b}}.

\bibitem[Chen and He(2021)]{chen2021exploring}
Xinlei Chen and Kaiming He.
\newblock Exploring simple siamese representation learning.
\newblock In \emph{Proceedings of the IEEE/CVF conference on computer vision and pattern recognition}, pages 15750--15758, 2021.

\bibitem[Chen et~al.(2016)Chen, Duan, Houthooft, Schulman, Sutskever, and Abbeel]{chen2016infogan}
Xi Chen, Yan Duan, Rein Houthooft, John Schulman, Ilya Sutskever, and Pieter Abbeel.
\newblock Infogan: Interpretable representation learning by information maximizing generative adversarial nets.
\newblock \emph{Advances in neural information processing systems}, 29, 2016.

\bibitem[Chen et~al.(2020{\natexlab{c}})Chen, Fan, Girshick, and He]{DBLP:journals/corr/abs-2003-04297}
Xinlei Chen, Haoqi Fan, Ross~B. Girshick, and Kaiming He.
\newblock Improved baselines with momentum contrastive learning.
\newblock \emph{CoRR}, abs/2003.04297, 2020{\natexlab{c}}.

\bibitem[Chen* et~al.(2021)Chen*, Xie*, and He]{chen2021mocov3}
Xinlei Chen*, Saining Xie*, and Kaiming He.
\newblock An empirical study of training self-supervised vision transformers.
\newblock \emph{arXiv preprint arXiv:2104.02057}, 2021.

\bibitem[Chen and Wang(2022)]{chen2022transformers}
Yinbo Chen and Xiaolong Wang.
\newblock Transformers as meta-learners for implicit neural representations, 2022.

\bibitem[Cimpoi et~al.(2014)Cimpoi, Maji, Kokkinos, Mohamed, , and Vedaldi]{cimpoi14describing}
M. Cimpoi, S. Maji, I. Kokkinos, S. Mohamed, , and A. Vedaldi.
\newblock Describing textures in the wild.
\newblock In \emph{Proceedings of the {IEEE} Conf. on Computer Vision and Pattern Recognition ({CVPR})}, 2014.

\bibitem[Deng et~al.(2009)Deng, Dong, Socher, Li, Li, and Fei-Fei]{imagenet}
Jia Deng, Wei Dong, Richard Socher, Li-Jia Li, Kai Li, and Li Fei-Fei.
\newblock Imagenet: A large-scale hierarchical image database.
\newblock In \emph{2009 IEEE Conference on Computer Vision and Pattern Recognition}, pages 248--255, 2009.

\bibitem[Donahue and Simonyan(2019)]{donahue2019large}
Jeff Donahue and Karen Simonyan.
\newblock Large scale adversarial representation learning.
\newblock \emph{Advances in neural information processing systems}, 32, 2019.

\bibitem[Donahue et~al.(2016)Donahue, Kr{\"a}henb{\"u}hl, and Darrell]{donahue2016adversarial}
Jeff Donahue, Philipp Kr{\"a}henb{\"u}hl, and Trevor Darrell.
\newblock Adversarial feature learning.
\newblock \emph{arXiv preprint arXiv:1605.09782}, 2016.

\bibitem[Dosovitskiy et~al.(2020)Dosovitskiy, Beyer, Kolesnikov, Weissenborn, Zhai, Unterthiner, Dehghani, Minderer, Heigold, Gelly, et~al.]{dosovitskiy2020image}
Alexey Dosovitskiy, Lucas Beyer, Alexander Kolesnikov, Dirk Weissenborn, Xiaohua Zhai, Thomas Unterthiner, Mostafa Dehghani, Matthias Minderer, Georg Heigold, Sylvain Gelly, et~al.
\newblock An image is worth 16x16 words: Transformers for image recognition at scale.
\newblock \emph{arXiv preprint arXiv:2010.11929}, 2020.

\bibitem[Dumoulin et~al.(2016)Dumoulin, Belghazi, Poole, Mastropietro, Lamb, Arjovsky, and Courville]{dumoulin2016adversarially}
Vincent Dumoulin, Ishmael Belghazi, Ben Poole, Olivier Mastropietro, Alex Lamb, Martin Arjovsky, and Aaron Courville.
\newblock Adversarially learned inference.
\newblock \emph{arXiv preprint arXiv:1606.00704}, 2016.

\bibitem[Dupont et~al.(2021)Dupont, Goli{\'n}ski, Alizadeh, Teh, and Doucet]{dupont2021coin}
Emilien Dupont, Adam Goli{\'n}ski, Milad Alizadeh, Yee~Whye Teh, and Arnaud Doucet.
\newblock Coin: Compression with implicit neural representations.
\newblock \emph{arXiv preprint arXiv:2103.03123}, 2021.

\bibitem[Dupont et~al.(2022)Dupont, Loya, Alizadeh, Golinski, Teh, and Doucet]{dupont2022coin++}
Emilien Dupont, Hrushikesh Loya, Milad Alizadeh, Adam Golinski, Yee~Whye Teh, and Arnaud Doucet.
\newblock Coin++: Data agnostic neural compression.
\newblock \emph{arXiv preprint arXiv:2201.12904}, 1\penalty0 (2):\penalty0 4, 2022.

\bibitem[El~Banani et~al.(2024)El~Banani, Raj, Maninis, Kar, Li, Rubinstein, Sun, Guibas, Johnson, and Jampani]{El_Banani_2024_CVPR}
Mohamed El~Banani, Amit Raj, Kevis-Kokitsi Maninis, Abhishek Kar, Yuanzhen Li, Michael Rubinstein, Deqing Sun, Leonidas Guibas, Justin Johnson, and Varun Jampani.
\newblock Probing the 3d awareness of visual foundation models.
\newblock In \emph{Proceedings of the IEEE/CVF Conference on Computer Vision and Pattern Recognition (CVPR)}, pages 21795--21806, 2024.

\bibitem[Gadre et~al.(2023)Gadre, Ilharco, Fang, Hayase, Smyrnis, Nguyen, Marten, Wortsman, Ghosh, Zhang, Orgad, Entezari, Daras, Pratt, Ramanujan, Bitton, Marathe, Mussmann, Vencu, Cherti, Krishna, Koh, Saukh, Ratner, Song, Hajishirzi, Farhadi, Beaumont, Oh, Dimakis, Jitsev, Carmon, Shankar, and Schmidt]{datacomp}
Samir~Yitzhak Gadre, Gabriel Ilharco, Alex Fang, Jonathan Hayase, Georgios Smyrnis, Thao Nguyen, Ryan Marten, Mitchell Wortsman, Dhruba Ghosh, Jieyu Zhang, Eyal Orgad, Rahim Entezari, Giannis Daras, Sarah Pratt, Vivek Ramanujan, Yonatan Bitton, Kalyani Marathe, Stephen Mussmann, Richard Vencu, Mehdi Cherti, Ranjay Krishna, Pang~Wei Koh, Olga Saukh, Alexander Ratner, Shuran Song, Hannaneh Hajishirzi, Ali Farhadi, Romain Beaumont, Sewoong Oh, Alex Dimakis, Jenia Jitsev, Yair Carmon, Vaishaal Shankar, and Ludwig Schmidt.
\newblock Datacomp: In search of the next generation of multimodal datasets.
\newblock \emph{arXiv preprint arXiv:2304.14108}, 2023.

\bibitem[Girish et~al.(2023)Girish, Shrivastava, and Gupta]{girish2023shacirascalablehashgridcompression}
Sharath Girish, Abhinav Shrivastava, and Kamal Gupta.
\newblock Shacira: Scalable hash-grid compression for implicit neural representations, 2023.

\bibitem[Gregor et~al.(2016)Gregor, Besse, Rezende, Danihelka, and Wierstra]{gregor2016conceptualcompression}
Karol Gregor, Frederic Besse, Danilo~Jimenez Rezende, Ivo Danihelka, and Daan Wierstra.
\newblock Towards conceptual compression, 2016.

\bibitem[Grill et~al.(2020)Grill, Strub, Altch{\'{e}}, Tallec, Richemond, Buchatskaya, Doersch, Pires, Guo, Azar, Piot, Kavukcuoglu, Munos, and Valko]{DBLP:journals/corr/abs-2006-07733}
Jean{-}Bastien Grill, Florian Strub, Florent Altch{\'{e}}, Corentin Tallec, Pierre~H. Richemond, Elena Buchatskaya, Carl Doersch, Bernardo~{\'{A}}vila Pires, Zhaohan~Daniel Guo, Mohammad~Gheshlaghi Azar, Bilal Piot, Koray Kavukcuoglu, R{\'{e}}mi Munos, and Michal Valko.
\newblock Bootstrap your own latent: {A} new approach to self-supervised learning.
\newblock \emph{CoRR}, abs/2006.07733, 2020.

\bibitem[Gwilliam et~al.(2025)Gwilliam, Zhang, Padmanabhan, Du, and Shrivastava]{gwilliam2025design}
Matthew Gwilliam, Roy Zhang, Namitha Padmanabhan, Hongyang Du, and Abhinav Shrivastava.
\newblock How to design and train your implicit neural representation for video compression.
\newblock \emph{arXiv preprint arXiv:2506.24127}, 2025.

\bibitem[Haydarov et~al.(2024)Haydarov, Muhamed, Shen, Lazarevic, Skorokhodov, Galappaththige, and Elhoseiny]{Haydarov_2024_CVPR}
Kilichbek Haydarov, Aashiq Muhamed, Xiaoqian Shen, Jovana Lazarevic, Ivan Skorokhodov, Chamuditha~Jayanga Galappaththige, and Mohamed Elhoseiny.
\newblock Adversarial text to continuous image generation.
\newblock In \emph{Proceedings of the IEEE/CVF Conference on Computer Vision and Pattern Recognition (CVPR)}, pages 6316--6326, 2024.

\bibitem[He et~al.(2023)He, Yang, Wang, Wu, Chen, Huang, Ren, Lim, and Shrivastava]{He_2023_CVPR}
Bo He, Xitong Yang, Hanyu Wang, Zuxuan Wu, Hao Chen, Shuaiyi Huang, Yixuan Ren, Ser-Nam Lim, and Abhinav Shrivastava.
\newblock Towards scalable neural representation for diverse videos.
\newblock In \emph{Proceedings of the IEEE/CVF Conference on Computer Vision and Pattern Recognition (CVPR)}, pages 6132--6142, 2023.

\bibitem[He et~al.(2021)He, Chen, Xie, Li, Doll{\'{a}}r, and Girshick]{DBLP:journals/corr/abs-2111-06377}
Kaiming He, Xinlei Chen, Saining Xie, Yanghao Li, Piotr Doll{\'{a}}r, and Ross~B. Girshick.
\newblock Masked autoencoders are scalable vision learners.
\newblock \emph{CoRR}, abs/2111.06377, 2021.

\bibitem[Heinrich et~al.(2024)Heinrich, Ranzinger, Hongxu, Yin, Lu, Kautz, Tao, Catanzaro, and Molchanov]{heinrich2025radiov25improvedbaselinesagglomerative}
Greg Heinrich, Mike Ranzinger, Hongxu, Yin, Yao Lu, Jan Kautz, Andrew Tao, Bryan Catanzaro, and Pavlo Molchanov.
\newblock Radiov2.5: Improved baselines for agglomerative vision foundation models, 2024.

\bibitem[Hinton et~al.(2015)Hinton, Vinyals, and Dean]{hinton2015distilling}
Geoffrey Hinton, Oriol Vinyals, and Jeff Dean.
\newblock Distilling the knowledge in a neural network.
\newblock \emph{arXiv preprint arXiv:1503.02531}, 2015.

\bibitem[Howard()]{imagenette}
Jeremy Howard.
\newblock Imagenette.

\bibitem[Huang et~al.(2022)Huang, Jin, Lu, Hou, Cheng, Fu, Shen, and Feng]{huang2022contrastive}
Zhicheng Huang, Xiaojie Jin, Chengze Lu, Qibin Hou, Ming-Ming Cheng, Dongmei Fu, Xiaohui Shen, and Jiashi Feng.
\newblock Contrastive masked autoencoders are stronger vision learners, 2022.

\bibitem[Kim et~al.(2022{\natexlab{a}})Kim, Lee, Kim, Cho, and Han]{kim2022generalizable}
Chiheon Kim, Doyup Lee, Saehoon Kim, Minsu Cho, and Wook-Shin Han.
\newblock Generalizable implicit neural representations via instance pattern composers.
\newblock \emph{arXiv preprint arXiv:2211.13223}, 2022{\natexlab{a}}.

\bibitem[Kim et~al.(2023)Kim, Bauer, Theis, Schwarz, and Dupont]{kim2023c3highperformancelowcomplexityneural}
Hyunjik Kim, Matthias Bauer, Lucas Theis, Jonathan~Richard Schwarz, and Emilien Dupont.
\newblock C3: High-performance and low-complexity neural compression from a single image or video, 2023.

\bibitem[Kim et~al.(2024)Kim, Lee, and Kang]{Kim_2024}
Jina Kim, Jihoo Lee, and Je-Won Kang.
\newblock \emph{SNeRV: Spectra-Preserving Neural Representation for Video}, page 332–348.
\newblock Springer Nature Switzerland, 2024.

\bibitem[Kim et~al.(2022{\natexlab{b}})Kim, Yu, Lee, and Shin]{kim2022scalable}
Subin Kim, Sihyun Yu, Jaeho Lee, and Jinwoo Shin.
\newblock Scalable neural video representations with learnable positional features.
\newblock In \emph{Advances in Neural Information Processing Systems}, 2022{\natexlab{b}}.

\bibitem[Krause et~al.(2013)Krause, Stark, Deng, and Fei-Fei]{KrauseStarkDengFei-Fei_3DRR2013}
Jonathan Krause, Michael Stark, Jia Deng, and Li Fei-Fei.
\newblock 3d object representations for fine-grained categorization.
\newblock In \emph{4th International IEEE Workshop on 3D Representation and Recognition (3dRR-13)}, Sydney, Australia, 2013.

\bibitem[Kwan et~al.(2023)Kwan, Gao, Zhang, Gower, and Bull]{kwan2023hinerv}
Ho~Man Kwan, Ge Gao, Fan Zhang, Andrew Gower, and David Bull.
\newblock Hinerv: Video compression with hierarchical encoding-based neural representation.
\newblock In \emph{Advances in Neural Information Processing Systems}, pages 72692--72704. Curran Associates, Inc., 2023.

\bibitem[Kwan et~al.(2024)Kwan, Gao, Zhang, Gower, and Bull]{kwan2024nvrcneuralvideorepresentation}
Ho~Man Kwan, Ge Gao, Fan Zhang, Andrew Gower, and David Bull.
\newblock Nvrc: Neural video representation compression, 2024.

\bibitem[Ladune et~al.(2023)Ladune, Philippe, Henry, Clare, and Leguay]{ladune2023coolchiccoordinatebasedlowcomplexity}
Théo Ladune, Pierrick Philippe, Félix Henry, Gordon Clare, and Thomas Leguay.
\newblock Cool-chic: Coordinate-based low complexity hierarchical image codec, 2023.

\bibitem[Lee et~al.(2023{\natexlab{a}})Lee, Kim, Cho, and Han]{lee2023localityawaregeneralizableimplicitneural}
Doyup Lee, Chiheon Kim, Minsu Cho, and Wook-Shin Han.
\newblock Locality-aware generalizable implicit neural representation, 2023{\natexlab{a}}.

\bibitem[Lee et~al.(2023{\natexlab{b}})Lee, Rho, Ko, and Park]{Lee_2023}
Joo~Chan Lee, Daniel Rho, Jong~Hwan Ko, and Eunbyung Park.
\newblock Ffnerv: Flow-guided frame-wise neural representations for videos.
\newblock In \emph{Proceedings of the 31st ACM International Conference on Multimedia}, page 7859–7870. ACM, 2023{\natexlab{b}}.

\bibitem[Lee et~al.(2021)Lee, Arnab, Guadarrama, Canny, and Fischer]{lee2021compressivevisualrepresentations}
Kuang-Huei Lee, Anurag Arnab, Sergio Guadarrama, John Canny, and Ian Fischer.
\newblock Compressive visual representations, 2021.

\bibitem[Li et~al.(2023)Li, Prabhudesai, Duggal, Brown, and Pathak]{li2023your}
Alexander~Cong Li, Mihir Prabhudesai, Shivam Duggal, Ellis~Langham Brown, and Deepak Pathak.
\newblock Your diffusion model is secretly a zero-shot classifier.
\newblock In \emph{ICML 2023 Workshop on Structured Probabilistic Inference {\&} Generative Modeling}, 2023.

\bibitem[Li et~al.(2022{\natexlab{a}})Li, Yang, Zhang, Gao, Xiao, Dai, Yuan, and Gao]{li2022efficient}
Chunyuan Li, Jianwei Yang, Pengchuan Zhang, Mei Gao, Bin Xiao, Xiyang Dai, Lu Yuan, and Jianfeng Gao.
\newblock Efficient self-supervised vision transformers for representation learning, 2022{\natexlab{a}}.

\bibitem[Li et~al.(2022{\natexlab{b}})Li, Chang, Mishra, Zhang, Katabi, and Krishnan]{li2022mage}
Tianhong Li, Huiwen Chang, Shlok~Kumar Mishra, Han Zhang, Dina Katabi, and Dilip Krishnan.
\newblock Mage: Masked generative encoder to unify representation learning and image synthesis, 2022{\natexlab{b}}.

\bibitem[Li et~al.(2022{\natexlab{c}})Li, Wang, Pi, Xu, Mei, and Liu]{li2022enervexpediteneuralvideo}
Zizhang Li, Mengmeng Wang, Huaijin Pi, Kechun Xu, Jianbiao Mei, and Yong Liu.
\newblock E-nerv: Expedite neural video representation with disentangled spatial-temporal context, 2022{\natexlab{c}}.

\bibitem[Liu and Nocedal(1989)]{liu1989limited}
Dong~C Liu and Jorge Nocedal.
\newblock On the limited memory bfgs method for large scale optimization.
\newblock \emph{Mathematical programming}, 45\penalty0 (1):\penalty0 503--528, 1989.

\bibitem[Liu et~al.(2023)Liu, Li, Wu, and Lee]{liu2023visual}
Haotian Liu, Chunyuan Li, Qingyang Wu, and Yong~Jae Lee.
\newblock Visual instruction tuning.
\newblock \emph{Advances in neural information processing systems}, 36:\penalty0 34892--34916, 2023.

\bibitem[Liu et~al.(2024)Liu, Li, Li, and Lee]{liu2024improved}
Haotian Liu, Chunyuan Li, Yuheng Li, and Yong~Jae Lee.
\newblock Improved baselines with visual instruction tuning.
\newblock In \emph{Proceedings of the IEEE/CVF conference on computer vision and pattern recognition}, pages 26296--26306, 2024.

\bibitem[Liu et~al.(2015)Liu, Luo, Wang, and Tang]{liu2015faceattributes}
Ziwei Liu, Ping Luo, Xiaogang Wang, and Xiaoou Tang.
\newblock Deep learning face attributes in the wild.
\newblock In \emph{Proceedings of International Conference on Computer Vision (ICCV)}, 2015.

\bibitem[Loshchilov and Hutter(2019)]{loshchilov2019decoupledweightdecayregularization}
Ilya Loshchilov and Frank Hutter.
\newblock Decoupled weight decay regularization, 2019.

\bibitem[Maiya et~al.(2023)Maiya, Girish, Ehrlich, Wang, Lee, Poirson, Wu, Wang, and Shrivastava]{maiya2023nirvana}
Shishira~R Maiya, Sharath Girish, Max Ehrlich, Hanyu Wang, Kwot~Sin Lee, Patrick Poirson, Pengxiang Wu, Chen Wang, and Abhinav Shrivastava.
\newblock Nirvana: Neural implicit representations of videos with adaptive networks and autoregressive patch-wise modeling.
\newblock In \emph{Proceedings of the IEEE/CVF Conference on Computer Vision and Pattern Recognition}, pages 14378--14387, 2023.

\bibitem[Maiya et~al.(2024)Maiya, Gupta, Gwilliam, Ehrlich, and Shrivastava]{maiya2024latent}
Shishira~R Maiya, Anubhav Gupta, Matthew Gwilliam, Max Ehrlich, and Abhinav Shrivastava.
\newblock Latent-inr: A flexible framework for implicit representations of videos with discriminative semantics.
\newblock In \emph{European Conference on Computer Vision}, pages 285--302. Springer, 2024.

\bibitem[Mildenhall et~al.(2020)Mildenhall, Srinivasan, Tancik, Barron, Ramamoorthi, and Ng]{mildenhall2020nerf}
Ben Mildenhall, Pratul~P. Srinivasan, Matthew Tancik, Jonathan~T. Barron, Ravi Ramamoorthi, and Ren Ng.
\newblock Nerf: Representing scenes as neural radiance fields for view synthesis, 2020.

\bibitem[Misra and Maaten(2020)]{misra2020self}
Ishan Misra and Laurens van~der Maaten.
\newblock Self-supervised learning of pretext-invariant representations.
\newblock In \emph{Proceedings of the IEEE/CVF Conference on Computer Vision and Pattern Recognition}, pages 6707--6717, 2020.

\bibitem[Mukhopadhyay et~al.(2024)Mukhopadhyay, Gwilliam, Yamaguchi, Agarwal, Padmanabhan, Swaminathan, Zhou, Ohya, and Shrivastava]{mukhopadhyay2024text}
Soumik Mukhopadhyay, Matthew Gwilliam, Yosuke Yamaguchi, Vatsal Agarwal, Namitha Padmanabhan, Archana Swaminathan, Tianyi Zhou, Jun Ohya, and Abhinav Shrivastava.
\newblock Do text-free diffusion models learn discriminative visual representations?
\newblock In \emph{European Conference on Computer Vision}, pages 253--272. Springer, 2024.

\bibitem[Müller et~al.(2022)Müller, Evans, Schied, and Keller]{M_ller_2022}
Thomas Müller, Alex Evans, Christoph Schied, and Alexander Keller.
\newblock Instant neural graphics primitives with a multiresolution hash encoding.
\newblock \emph{ACM Transactions on Graphics}, 41\penalty0 (4):\penalty0 1–15, 2022.

\bibitem[Nathan~Silberman and Fergus(2012)]{Silberman:ECCV12}
Pushmeet~Kohli Nathan~Silberman, Derek~Hoiem and Rob Fergus.
\newblock Indoor segmentation and support inference from rgbd images.
\newblock In \emph{ECCV}, 2012.

\bibitem[Nie et~al.(2020)Nie, Karras, Garg, Debnath, Patney, Patel, and Anandkumar]{nie2020semi}
Weili Nie, Tero Karras, Animesh Garg, Shoubhik Debnath, Anjul Patney, Ankit~B Patel, and Anima Anandkumar.
\newblock Semi-supervised stylegan for disentanglement learning.
\newblock In \emph{Proceedings of the 37th International Conference on Machine Learning}, pages 7360--7369, 2020.

\bibitem[Nilsback and Zisserman(2008)]{Nilsback08}
Maria-Elena Nilsback and Andrew Zisserman.
\newblock Automated flower classification over a large number of classes.
\newblock In \emph{Indian Conference on Computer Vision, Graphics and Image Processing}, 2008.

\bibitem[Noroozi and Favaro(2016)]{noroozi2016unsupervised}
Mehdi Noroozi and Paolo Favaro.
\newblock Unsupervised learning of visual representations by solving jigsaw puzzles.
\newblock In \emph{European conference on computer vision}, pages 69--84. Springer, 2016.

\bibitem[Oquab et~al.(2023)Oquab, Darcet, Moutakanni, Vo, Szafraniec, Khalidov, Fernandez, Haziza, Massa, El-Nouby, Assran, Ballas, Galuba, Howes, Huang, Li, Misra, Rabbat, Sharma, Synnaeve, Xu, Jegou, Mairal, Labatut, Joulin, and Bojanowski]{oquab2023dinov2}
Maxime Oquab, Timothée Darcet, Théo Moutakanni, Huy Vo, Marc Szafraniec, Vasil Khalidov, Pierre Fernandez, Daniel Haziza, Francisco Massa, Alaaeldin El-Nouby, Mahmoud Assran, Nicolas Ballas, Wojciech Galuba, Russell Howes, Po-Yao Huang, Shang-Wen Li, Ishan Misra, Michael Rabbat, Vasu Sharma, Gabriel Synnaeve, Hu Xu, Hervé Jegou, Julien Mairal, Patrick Labatut, Armand Joulin, and Piotr Bojanowski.
\newblock Dinov2: Learning robust visual features without supervision, 2023.

\bibitem[Pang et~al.(2022)Pang, Zhang, Li, Cai, and Lu]{pang2022unsupervised}
Bo Pang, Yifan Zhang, Yaoyi Li, Jia Cai, and Cewu Lu.
\newblock Unsupervised visual representation learning by synchronous momentum grouping, 2022.

\bibitem[Pathak et~al.(2016)Pathak, Krahenbuhl, Donahue, Darrell, and Efros]{pathak2016context}
Deepak Pathak, Philipp Krahenbuhl, Jeff Donahue, Trevor Darrell, and Alexei~A Efros.
\newblock Context encoders: Feature learning by inpainting.
\newblock In \emph{Proceedings of the IEEE conference on computer vision and pattern recognition}, pages 2536--2544, 2016.

\bibitem[Peebles and Xie(2023)]{peebles2023scalablediffusionmodelstransformers}
William Peebles and Saining Xie.
\newblock Scalable diffusion models with transformers, 2023.

\bibitem[Radford et~al.(2021)Radford, Kim, Hallacy, Ramesh, Goh, Agarwal, Sastry, Askell, Mishkin, Clark, et~al.]{radford2021learning}
Alec Radford, Jong~Wook Kim, Chris Hallacy, Aditya Ramesh, Gabriel Goh, Sandhini Agarwal, Girish Sastry, Amanda Askell, Pamela Mishkin, Jack Clark, et~al.
\newblock Learning transferable visual models from natural language supervision.
\newblock In \emph{International conference on machine learning}, pages 8748--8763. PmLR, 2021.

\bibitem[Ranzinger et~al.(2024)Ranzinger, Heinrich, Kautz, and Molchanov]{Ranzinger_2024_CVPR}
Mike Ranzinger, Greg Heinrich, Jan Kautz, and Pavlo Molchanov.
\newblock Am-radio: Agglomerative vision foundation model reduce all domains into one.
\newblock In \emph{Proceedings of the IEEE/CVF Conference on Computer Vision and Pattern Recognition (CVPR)}, pages 12490--12500, 2024.

\bibitem[Rombach et~al.(2022)Rombach, Blattmann, Lorenz, Esser, and Ommer]{rombach2022highresolutionimagesynthesislatent}
Robin Rombach, Andreas Blattmann, Dominik Lorenz, Patrick Esser, and Björn Ommer.
\newblock High-resolution image synthesis with latent diffusion models, 2022.

\bibitem[Russakovsky et~al.(2015)Russakovsky, Deng, Su, Krause, Satheesh, Ma, Huang, Karpathy, Khosla, Bernstein, Berg, and Fei-Fei]{imagenet15russakovsky}
Olga Russakovsky, Jia Deng, Hao Su, Jonathan Krause, Sanjeev Satheesh, Sean Ma, Zhiheng Huang, Andrej Karpathy, Aditya Khosla, Michael Bernstein, Alexander~C. Berg, and Li Fei-Fei.
\newblock {ImageNet Large Scale Visual Recognition Challenge}.
\newblock \emph{International Journal of Computer Vision (IJCV)}, 115\penalty0 (3):\penalty0 211--252, 2015.

\bibitem[Saethre et~al.(2024)Saethre, Azevedo, and Schroers]{Saethre_2024_CVPR}
Jens~Eirik Saethre, Roberto Azevedo, and Christopher Schroers.
\newblock Combining frame and gop embeddings for neural video representation.
\newblock In \emph{Proceedings of the IEEE/CVF Conference on Computer Vision and Pattern Recognition (CVPR)}, pages 9253--9263, 2024.

\bibitem[Saragadam et~al.(2023)Saragadam, LeJeune, Tan, Balakrishnan, Veeraraghavan, and Baraniuk]{saragadam2023wire}
Vishwanath Saragadam, Daniel LeJeune, Jasper Tan, Guha Balakrishnan, Ashok Veeraraghavan, and Richard~G. Baraniuk.
\newblock Wire: Wavelet implicit neural representations, 2023.

\bibitem[Schmidhuber(2006)]{schmidhuber2006computer}
J{\"u}rgen Schmidhuber.
\newblock A computer scientist's view of life, the universe, and everything.
\newblock In \emph{Foundations of computer science: Potential—theory—cognition}, pages 201--208. Springer, 2006.

\bibitem[Shazeer(2020)]{shazeer2020gluvariantsimprovetransformer}
Noam Shazeer.
\newblock Glu variants improve transformer, 2020.

\bibitem[Shi et~al.(2016)Shi, Caballero, Huszár, Totz, Aitken, Bishop, Rueckert, and Wang]{shi2016realtimesingleimagevideo}
Wenzhe Shi, Jose Caballero, Ferenc Huszár, Johannes Totz, Andrew~P. Aitken, Rob Bishop, Daniel Rueckert, and Zehan Wang.
\newblock Real-time single image and video super-resolution using an efficient sub-pixel convolutional neural network, 2016.

\bibitem[Sim{\'e}oni et~al.(2025)Sim{\'e}oni, Vo, Seitzer, Baldassarre, Oquab, Jose, Khalidov, Szafraniec, Yi, Ramamonjisoa, et~al.]{simeoni2025dinov3}
Oriane Sim{\'e}oni, Huy~V Vo, Maximilian Seitzer, Federico Baldassarre, Maxime Oquab, Cijo Jose, Vasil Khalidov, Marc Szafraniec, Seungeun Yi, Micha{\"e}l Ramamonjisoa, et~al.
\newblock Dinov3.
\newblock \emph{arXiv preprint arXiv:2508.10104}, 2025.

\bibitem[Sitzmann et~al.(2020)Sitzmann, Martel, Bergman, Lindell, and Wetzstein]{sitzmann2020implicit}
Vincent Sitzmann, Julien Martel, Alexander Bergman, David Lindell, and Gordon Wetzstein.
\newblock Implicit neural representations with periodic activation functions.
\newblock \emph{Advances in neural information processing systems}, 33:\penalty0 7462--7473, 2020.

\bibitem[Skorokhodov et~al.(2021)Skorokhodov, Ignatyev, and Elhoseiny]{skorokhodov2021adversarialgenerationcontinuousimages}
Ivan Skorokhodov, Savva Ignatyev, and Mohamed Elhoseiny.
\newblock Adversarial generation of continuous images, 2021.

\bibitem[Solomonoff(1964)]{solomonoff1964formal}
Ray~J Solomonoff.
\newblock A formal theory of inductive inference. part i.
\newblock \emph{Information and control}, 7\penalty0 (1):\penalty0 1--22, 1964.

\bibitem[Strümpler et~al.(2022)Strümpler, Postels, Yang, van Gool, and Tombari]{strümpler2022implicitneuralrepresentationsimage}
Yannick Strümpler, Janis Postels, Ren Yang, Luc van Gool, and Federico Tombari.
\newblock Implicit neural representations for image compression, 2022.

\bibitem[Su et~al.(2024)Su, Ahmed, Lu, Pan, Bo, and Liu]{su2024roformer}
Jianlin Su, Murtadha Ahmed, Yu Lu, Shengfeng Pan, Wen Bo, and Yunfeng Liu.
\newblock Roformer: Enhanced transformer with rotary position embedding.
\newblock \emph{Neurocomputing}, 568:\penalty0 127063, 2024.

\bibitem[Sun et~al.(2024)Sun, Wang, Yu, Cui, Zhang, Zhang, and Wang]{sun2024eva}
Quan Sun, Jinsheng Wang, Qiying Yu, Yufeng Cui, Fan Zhang, Xiaosong Zhang, and Xinlong Wang.
\newblock Eva-clip-18b: Scaling clip to 18 billion parameters.
\newblock \emph{arXiv preprint arXiv:2402.04252}, 2024.

\bibitem[Tancik et~al.(2020)Tancik, Srinivasan, Mildenhall, Fridovich-Keil, Raghavan, Singhal, Ramamoorthi, Barron, and Ng]{tancik2020fourfeat}
Matthew Tancik, Pratul~P. Srinivasan, Ben Mildenhall, Sara Fridovich-Keil, Nithin Raghavan, Utkarsh Singhal, Ravi Ramamoorthi, Jonathan~T. Barron, and Ren Ng.
\newblock Fourier features let networks learn high frequency functions in low dimensional domains.
\newblock \emph{NeurIPS}, 2020.

\bibitem[Tomasev et~al.(2022)Tomasev, Bica, McWilliams, Buesing, Pascanu, Blundell, and Mitrovic]{tomasev2022pushing}
Nenad Tomasev, Ioana Bica, Brian McWilliams, Lars Buesing, Razvan Pascanu, Charles Blundell, and Jovana Mitrovic.
\newblock Pushing the limits of self-supervised resnets: Can we outperform supervised learning without labels on imagenet?, 2022.

\bibitem[Tschannen et~al.(2025)Tschannen, Gritsenko, Wang, Naeem, Alabdulmohsin, Parthasarathy, Evans, Beyer, Xia, Mustafa, Hénaff, Harmsen, Steiner, and Zhai]{tschannen2025siglip2multilingualvisionlanguage}
Michael Tschannen, Alexey Gritsenko, Xiao Wang, Muhammad~Ferjad Naeem, Ibrahim Alabdulmohsin, Nikhil Parthasarathy, Talfan Evans, Lucas Beyer, Ye Xia, Basil Mustafa, Olivier Hénaff, Jeremiah Harmsen, Andreas Steiner, and Xiaohua Zhai.
\newblock Siglip 2: Multilingual vision-language encoders with improved semantic understanding, localization, and dense features, 2025.

\bibitem[Wah et~al.(2011)Wah, Branson, Welinder, Perona, and Belongie]{WahCUB_200_2011}
C. Wah, S. Branson, P. Welinder, P. Perona, and S. Belongie.
\newblock {The Caltech-UCSD Birds-200-2011 Dataset}.
\newblock Technical Report CNS-TR-2011-001, California Institute of Technology, 2011.

\bibitem[Wang et~al.(2024)Wang, Bai, Tan, Wang, Fan, Bai, Chen, Liu, Wang, Ge, et~al.]{wang2024qwen2}
Peng Wang, Shuai Bai, Sinan Tan, Shijie Wang, Zhihao Fan, Jinze Bai, Keqin Chen, Xuejing Liu, Jialin Wang, Wenbin Ge, et~al.
\newblock Qwen2-vl: Enhancing vision-language model's perception of the world at any resolution.
\newblock \emph{arXiv preprint arXiv:2409.12191}, 2024.

\bibitem[Wang et~al.(2004)Wang, Bovik, Sheikh, and Simoncelli]{wang2004image}
Zhou Wang, Alan~C Bovik, Hamid~R Sheikh, and Eero~P Simoncelli.
\newblock Image quality assessment: from error visibility to structural similarity.
\newblock \emph{IEEE transactions on image processing}, 13\penalty0 (4):\penalty0 600--612, 2004.

\bibitem[Wu et~al.(2024)Wu, Quan, He, Lai, Li, Yu, Lin, and Yang]{wu2024qs}
Chang Wu, Guancheng Quan, Gang He, Xin-Quan Lai, Yunsong Li, Wenxin Yu, Xianmeng Lin, and Cheng Yang.
\newblock Qs-nerv: Real-time quality-scalable decoding with neural representation for videos.
\newblock In \emph{Proceedings of the 32nd ACM International Conference on Multimedia}, pages 2584--2592, 2024.

\bibitem[Xiang et~al.(2023)Xiang, Yang, Huang, and Wang]{Xiang_2023_ICCV}
Weilai Xiang, Hongyu Yang, Di Huang, and Yunhong Wang.
\newblock Denoising diffusion autoencoders are unified self-supervised learners.
\newblock In \emph{Proceedings of the IEEE/CVF International Conference on Computer Vision (ICCV)}, pages 15802--15812, 2023.

\bibitem[Xu et~al.(2022)Xu, Wang, Jiang, Fan, and Wang]{xu2022signal}
Dejia Xu, Peihao Wang, Yifan Jiang, Zhiwen Fan, and Zhangyang Wang.
\newblock Signal processing for implicit neural representations, 2022.

\bibitem[Xu et~al.(2024)Xu, Feng, Qin, Ge, Peng, and Wang]{xu2024vqnervvectorquantizedneural}
Yunjie Xu, Xiang Feng, Feiwei Qin, Ruiquan Ge, Yong Peng, and Changmiao Wang.
\newblock Vq-nerv: A vector quantized neural representation for videos, 2024.

\bibitem[Xue et~al.(2024)Xue, Shu, Awadalla, Wang, Yan, Purushwalkam, Zhou, Prabhu, Dai, Ryoo, et~al.]{xue2024xgen}
Le Xue, Manli Shu, Anas Awadalla, Jun Wang, An Yan, Senthil Purushwalkam, Honglu Zhou, Viraj Prabhu, Yutong Dai, Michael~S Ryoo, et~al.
\newblock xgen-mm (blip-3): A family of open large multimodal models.
\newblock \emph{arXiv preprint arXiv:2408.08872}, 2024.

\bibitem[Yan et~al.(2024)Yan, Ke, Zhou, Qiu, Shi, and Jiang]{Yan_2024_CVPR}
Hao Yan, Zhihui Ke, Xiaobo Zhou, Tie Qiu, Xidong Shi, and Dadong Jiang.
\newblock Ds-nerv: Implicit neural video representation with decomposed static and dynamic codes.
\newblock In \emph{Proceedings of the IEEE/CVF Conference on Computer Vision and Pattern Recognition (CVPR)}, pages 23019--23029, 2024.

\bibitem[Yu et~al.(2015)Yu, Seff, Zhang, Song, Funkhouser, and Xiao]{yu2015lsun}
Fisher Yu, Ari Seff, Yinda Zhang, Shuran Song, Thomas Funkhouser, and Jianxiong Xiao.
\newblock Lsun: Construction of a large-scale image dataset using deep learning with humans in the loop.
\newblock \emph{arXiv preprint arXiv:1506.03365}, 2015.

\bibitem[Yu et~al.(2022)Yu, Tack, Mo, Kim, Kim, Ha, and Shin]{yu2022generatingvideosdynamicsawareimplicit}
Sihyun Yu, Jihoon Tack, Sangwoo Mo, Hyunsu Kim, Junho Kim, Jung-Woo Ha, and Jinwoo Shin.
\newblock Generating videos with dynamics-aware implicit generative adversarial networks, 2022.

\bibitem[Zbontar et~al.(2021)Zbontar, Jing, Misra, LeCun, and Deny]{pmlr-v139-zbontar21a}
Jure Zbontar, Li Jing, Ishan Misra, Yann LeCun, and Stephane Deny.
\newblock Barlow twins: Self-supervised learning via redundancy reduction.
\newblock In \emph{Proceedings of the 38th International Conference on Machine Learning}, pages 12310--12320. PMLR, 2021.

\bibitem[Zhai et~al.(2023)Zhai, Mustafa, Kolesnikov, and Beyer]{zhai2023sigmoid}
Xiaohua Zhai, Basil Mustafa, Alexander Kolesnikov, and Lucas Beyer.
\newblock Sigmoid loss for language image pre-training.
\newblock In \emph{Proceedings of the IEEE/CVF international conference on computer vision}, pages 11975--11986, 2023.

\bibitem[Zhang et~al.(2016)Zhang, Isola, and Efros]{zhang2016colorful}
Richard Zhang, Phillip Isola, and Alexei~A Efros.
\newblock Colorful image colorization.
\newblock In \emph{European conference on computer vision}, pages 649--666. Springer, 2016.

\bibitem[Zhang et~al.(2018)Zhang, Isola, Efros, Shechtman, and Wang]{zhang2018perceptual}
Richard Zhang, Phillip Isola, Alexei~A Efros, Eli Shechtman, and Oliver Wang.
\newblock The unreasonable effectiveness of deep features as a perceptual metric.
\newblock In \emph{CVPR}, 2018.

\bibitem[Zhang et~al.(2024{\natexlab{a}})Zhang, Liu, Gu, Cai, Wang, Bu, and Wang]{zhang2024attention}
Shuyi Zhang, Ke Liu, Jingjun Gu, Xiaoxu Cai, Zhihua Wang, Jiajun Bu, and Haishuai Wang.
\newblock Attention beats linear for fast implicit neural representation generation.
\newblock In \emph{European Conference on Computer Vision}, pages 1--18. Springer, 2024{\natexlab{a}}.

\bibitem[Zhang et~al.(2024{\natexlab{b}})Zhang, Yang, He, Ge, Xu, Wang, Qin, and Zhang]{Zhang_2024_CVPR}
Xinjie Zhang, Ren Yang, Dailan He, Xingtong Ge, Tongda Xu, Yan Wang, Hongwei Qin, and Jun Zhang.
\newblock Boosting neural representations for videos with a conditional decoder.
\newblock In \emph{Proceedings of the IEEE/CVF Conference on Computer Vision and Pattern Recognition (CVPR)}, pages 2556--2566, 2024{\natexlab{b}}.

\bibitem[Zhang et~al.(2021)Zhang, van Rozendaal, Brehmer, Nagel, and Cohen]{zhang2021implicitneuralvideocompression}
Yunfan Zhang, Ties van Rozendaal, Johann Brehmer, Markus Nagel, and Taco Cohen.
\newblock Implicit neural video compression, 2021.

\bibitem[Zhao et~al.(2023)Zhao, Asif, and Ma]{Zhao_2023_CVPR}
Qi Zhao, M.~Salman Asif, and Zhan Ma.
\newblock Dnerv: Modeling inherent dynamics via difference neural representation for videos.
\newblock In \emph{Proceedings of the IEEE/CVF Conference on Computer Vision and Pattern Recognition (CVPR)}, pages 2031--2040, 2023.

\bibitem[Zhao et~al.(2024)Zhao, Asif, and Ma]{Zhao_2024_CVPR}
Qi Zhao, M.~Salman Asif, and Zhan Ma.
\newblock Pnerv: Enhancing spatial consistency via pyramidal neural representation for videos.
\newblock In \emph{Proceedings of the IEEE/CVF Conference on Computer Vision and Pattern Recognition (CVPR)}, pages 19103--19112, 2024.

\bibitem[Zheng et~al.(2025)Zheng, Ma, Tong, and Xie]{zheng2025diffusiontransformersrepresentationautoencoders}
Boyang Zheng, Nanye Ma, Shengbang Tong, and Saining Xie.
\newblock Diffusion transformers with representation autoencoders, 2025.

\bibitem[Zhou et~al.(2017)Zhou, Zhao, Puig, Fidler, Barriuso, and Torralba]{zhou2017scene}
Bolei Zhou, Hang Zhao, Xavier Puig, Sanja Fidler, Adela Barriuso, and Antonio Torralba.
\newblock Scene parsing through ade20k dataset.
\newblock In \emph{Proceedings of the IEEE conference on computer vision and pattern recognition}, pages 633--641, 2017.

\bibitem[Zhou et~al.(2022{\natexlab{a}})Zhou, Wei, Wang, Shen, Xie, Yuille, and Kong]{zhou2022ibot}
Jinghao Zhou, Chen Wei, Huiyu Wang, Wei Shen, Cihang Xie, Alan Yuille, and Tao Kong.
\newblock ibot: Image bert pre-training with online tokenizer, 2022{\natexlab{a}}.

\bibitem[Zhou et~al.(2022{\natexlab{b}})Zhou, Zhou, Si, Yu, Ng, and Yan]{zhou2022mugs}
Pan Zhou, Yichen Zhou, Chenyang Si, Weihao Yu, Teck~Khim Ng, and Shuicheng Yan.
\newblock Mugs: A multi-granular self-supervised learning framework, 2022{\natexlab{b}}.

\end{thebibliography}
\end{document}